%% file: main.tex
\documentclass[11pt]{article}

\usepackage[margin=1in]{geometry}
\usepackage{booktabs}
\usepackage{array}
\usepackage{graphicx}
\usepackage{xcolor}
\usepackage{hyperref}
\usepackage[numbers,square,sort&compress]{natbib}
\usepackage{enumitem}
\usepackage{caption}
\usepackage{subcaption}
\usepackage{amsmath}
\usepackage{amssymb}
\usepackage{float}

\newcommand{\FloatBarrier}{}
\hypersetup{
  colorlinks=true,
  linkcolor=blue,
  citecolor=blue,
  urlcolor=blue
}

\title{\textbf{Do Larger Models Really Win in Drug Discovery?}\\
\large A Benchmark Assessment of Model Scaling in AI Driven Molecular Property and Activity Prediction}

\author{
  Jinjiang Guo$^{1,*}$ and Sheng Ding$^{1,2,*}$\\[0.5em]
  $^{1}$Global Health Drug Discovery Institute, Beijing 100084, China\\
  $^{2}$School of Pharmaceutical Sciences, Tsinghua University, Beijing 100084, China\\[0.5em]
  $^{*}$Correspondence: \texttt{jinjiang.guo@ghddi.org}; \texttt{sheng.ding@ghddi.org}
}

\date{}

\begin{document}
\maketitle

\begin{abstract}
The rapid growth of molecular foundation models and large language models (LLMs) has encouraged a scale centred view of AI in drug discovery, in which larger pretrained models are expected to supersede compact cheminformatics models based on classical machine learning (classical ML) and graph neural networks (GNNs) trained for individual tasks. We test this assumption across 26 endpoints grouped into ADME, toxicity and bioactivity classes, covering 165{,}541 endpoint level compound label records. The benchmark contains 78 endpoint and split entries, corresponding to 26 endpoints evaluated under three split protocols: random, Murcko scaffold and structure separated 5-fold cross validation (CV). Ordered from easiest to hardest, these splits approximate retrospective evaluation on a closed library, scaffold expansion in hit to lead, and library expansion on novel chemotypes. Each entry contributes two task and metric comparisons, giving 156 comparisons in total. Across these comparisons, classical ML provides the largest share of best performing entries (47.4\%), followed by pretrained molecular sequence models (28.8\%), GNNs (21.8\%) and LLM based SAR baselines (1.9\%). Classical ML dominates random split interpolation and remains the largest winner family overall. GNN and sequence models are competitive in selected harder split protocols under the primary optimal held-out readout, but their strict winner shares decrease under a fixed final-window readout, indicating that some of these gains depend on training settings and model selection. Paired bootstrap analyses indicate that small numerical differences between individual models should not be read as decisive victories. SAR knowledge from training folds improves many GPT5.5-SAR and Opus4.7-SAR metrics but does not make rule based reasoning a universal substitute for supervised predictors. Compact specialized models remain highly effective for molecular property and activity prediction. Larger models add value for SAR interpretation and reasoning in low data settings, but predictive performance depends on the fit among model, task and validation scenario, not on scale alone.
\end{abstract}


\section{Introduction}

AI driven drug discovery has entered a scale aware phase. Molecular graph neural networks (GNNs), pretrained SMILES transformers, molecular foundation models and general purpose large language models (LLMs) now coexist with classical machine learning (classical ML) models for quantitative structure activity relationship (QSAR) modelling, built from fingerprints and physicochemical descriptors. This expansion has broadened the modelling toolkit. Pretrained models can encode regularities across large chemical corpora. GNNs can learn task specific structural representations. LLMs can express medicinal chemistry reasoning through a flexible natural language interface. Yet this progress has also sharpened a practical question: do larger models reliably yield better molecular predictions?

For small molecule property and activity prediction, a universal scaling answer is unlikely. Many drug discovery datasets are small, noisy, imbalanced, endpoint specific and chemically clustered. A model pretrained on millions or billions of molecules may still be asked to predict a narrow assay readout from a few hundred to a few thousand labelled compounds. Conversely, a random forest or ExtraTrees model using extended connectivity fingerprints (ECFPs) or RDKit descriptors can align closely with local SAR patterns. Previous benchmarks and few shot molecular learning resources have similarly shown that learned representations do not automatically dominate classical molecular representations under data scarcity or distribution shift \citep{moleculenet2018,tdc2021,stanley2021fsmol,limits_rep_learning2023,why_deep_not_beat2023,admet_features2025}.

Here we turn this critique into a benchmark assessment of model scaling in molecular property and activity prediction. We do not ask whether any model family is globally superior. Instead, we ask where fingerprint classical ML models, GNNs, pretrained molecular sequence models and LLM-SAR reasoning baselines are useful, and where they fail. The central hypothesis is intentionally conservative:

\begin{quote}
Larger models can be useful, but scale alone may not explain predictive performance in drug discovery. Endpoint level performance may instead reflect model task fit: the alignment among molecular representation, inductive bias, training strategy, endpoint biology, data regime and validation setting.
\end{quote}

The models span several orders of magnitude in effective complexity: tree ensembles with hundreds of decision trees, GCN, GAT, GIN and Ligandformer style graph models with roughly $10^5$ to $10^6$ learned parameters, ChemBERTa and ChemBERTa2 checkpoints with 77M and 10M parameters, and MoLFormer with approximately 47M parameters pretrained on large SMILES corpora \citep{chemberta2020,chemberta22022,molformer2022}. The LLM-SAR baselines represent a different scaling regime: GPT5.5-SAR and Opus4.7-SAR use frontier LLM chemical reasoning to author and refine endpoint specific SAR rules, but the held out predictions are produced by rule application rather than by fine tuning or direct queries to a local parameterized predictor. This range supports four questions. Does parameter count predict endpoint level performance? Does pretraining corpus size predict winning? When do fingerprints, graphs, SMILES tokens or explicit SAR rules provide the most useful inductive bias? Where does LLM assisted SAR reasoning help, and where can it mislead?

\section{Benchmark Scope}

The benchmark covers 26 endpoint, label and source units and 78 split and task entries, with 165{,}541 endpoint level compound label records in total. To avoid conflating mechanistically different assays under a single ADMET label, we use three main endpoint classes: ADME related, toxicity related and bioactivity related endpoints. Endpoint subtypes are retained in Table~\ref{tab:endpoint-classes}.

The benchmark tasks are grouped as follows.

\begin{itemize}[leftmargin=*, itemsep=0.35em]
\item \textbf{ADME related endpoints.} This class contains six endpoint, label and source units and 18 split and task entries: Caco2 permeability, blood brain barrier penetration, CYP3A4 inhibition, PXR/NR1I2 activation, lipophilicity and aqueous solubility \citep{tdc2021,moleculenet2018,mendez2019bioassay}. These endpoints cover absorption, distribution, metabolism, drug interaction liability and physicochemical properties that shape developability. PXR/NR1I2 activation is placed here because PXR regulates xenobiotic response genes, drug metabolizing enzymes and transporters \citep{kliewer1998pxr}.

\item \textbf{Toxicity related endpoints.} This class contains 17 endpoint, label and source units and 51 split and task entries: AMES mutagenicity, DILI, two independently sourced KCNH2/hERG liability labels, twelve Tox21 nuclear receptor and stress response assays, and DRD2 receptor binding \citep{tdc2021,moleculenet2018,tox21_challenge2015,tox21_qsar2016,mendez2019bioassay}. KCNH2/hERG is treated as cardiotoxicity and safety pharmacology because it reflects cardiac \(I_{\mathrm{Kr}}\) inhibition risk \citep{sanguinetti1995herg}. The two KCNH2/hERG labels differ in provenance and experimental readout. The TDC hERG task is a small binary hERG blocker benchmark derived from curated blocker and non-blocker annotations \citep{tdc2021,wang2016herg}. The ChEMBL KCNH2/hERG task is a larger human KCNH2 IC$_{50}$ aggregation of hERG current or Kv11.1 inhibition assays, binarized here as active at pIC$_{50}\geq 6$ and inactive at pIC$_{50}\leq 5$, with intermediate potencies removed \citep{mendez2019bioassay}. DRD2 receptor binding is treated as an off target CNS safety liability endpoint because dopamine D2 receptor engagement is linked to extrapyramidal and prolactin related liabilities in antipsychotic pharmacology \citep{deGreef2011d2occupancy}.

\item \textbf{Bioactivity related endpoints.} This class contains 3 endpoint, label and source units and 9 split and task entries: EGFR kinase inhibition, anti-TB H37Rv cellular activity and antimalaria Pf. 3D7/Dd2 cellular activity. EGFR is treated as a target based bioactivity benchmark because it is a clinically validated oncology kinase target with extensive small molecule inhibitor SAR data~\citep{roskoski2019egfr,cohen2021kinase}. The primary anti-TB H37Rv dataset is a public core growth inhibition panel, with compounds treated as active at a minimum inhibitory concentration threshold of 1~$\mu$M. H37Rv is the canonical virulent laboratory reference strain for tuberculosis biology \citep{chitale2022h37rv}. Public anti-TB H37Rv records were assembled from open screening releases and literature datasets, including GSK/Tres Cantos open source TB leads, Rebollo-Lopez public compounds, the Lane and Ekins \textit{M. tuberculosis} machine learning dataset, and MLSMR follow up data \citep{ballell2013opensourceTB,rebollo2015opensourceTB,lane2018mtbml,eke2025mlsmrTB}. The primary antimalaria Pf. 3D7/Dd2 dataset uses public cellular activity records against \textit{P. falciparum} 3D7 and Dd2 parasites, with EC$_{50}\leq$100~nM as the active cutoff. The 3D7 strain is a drug sensitive laboratory reference, whereas Dd2 is a drug resistant line used to probe resistance liabilities, including resistance to chloroquine linked to mutated \textit{pfcrt} \citep{iwanaga2022pfresistance}. Restricted institutional infectious disease datasets are reported separately as supplementary restricted data analyses (Supplementary Note~5) and are not used in the primary family win denominator.
\end{itemize}

Table~\ref{tab:endpoint-classes} summarizes this endpoint class mapping and is used consistently in the result tables and figures. Table~\ref{tab:dataset-sizes} reports the size and positive rate of every endpoint, with provenance for each label source.

Across the unified endpoint panel, classical ML, GNN, sequence and LLM-SAR comparisons are summarized over 78 endpoint and split entries spanning structure separated, random and Murcko scaffold 5-fold cross validation. Classification entries include all four model families; ADME related regression entries include the families with available regression outputs.

\section{Model Families and Molecular Representations}
\label{sec:models}

\begin{figure}[!htbp]
\centering
\includegraphics[width=0.86\linewidth,height=0.45\textheight,keepaspectratio]{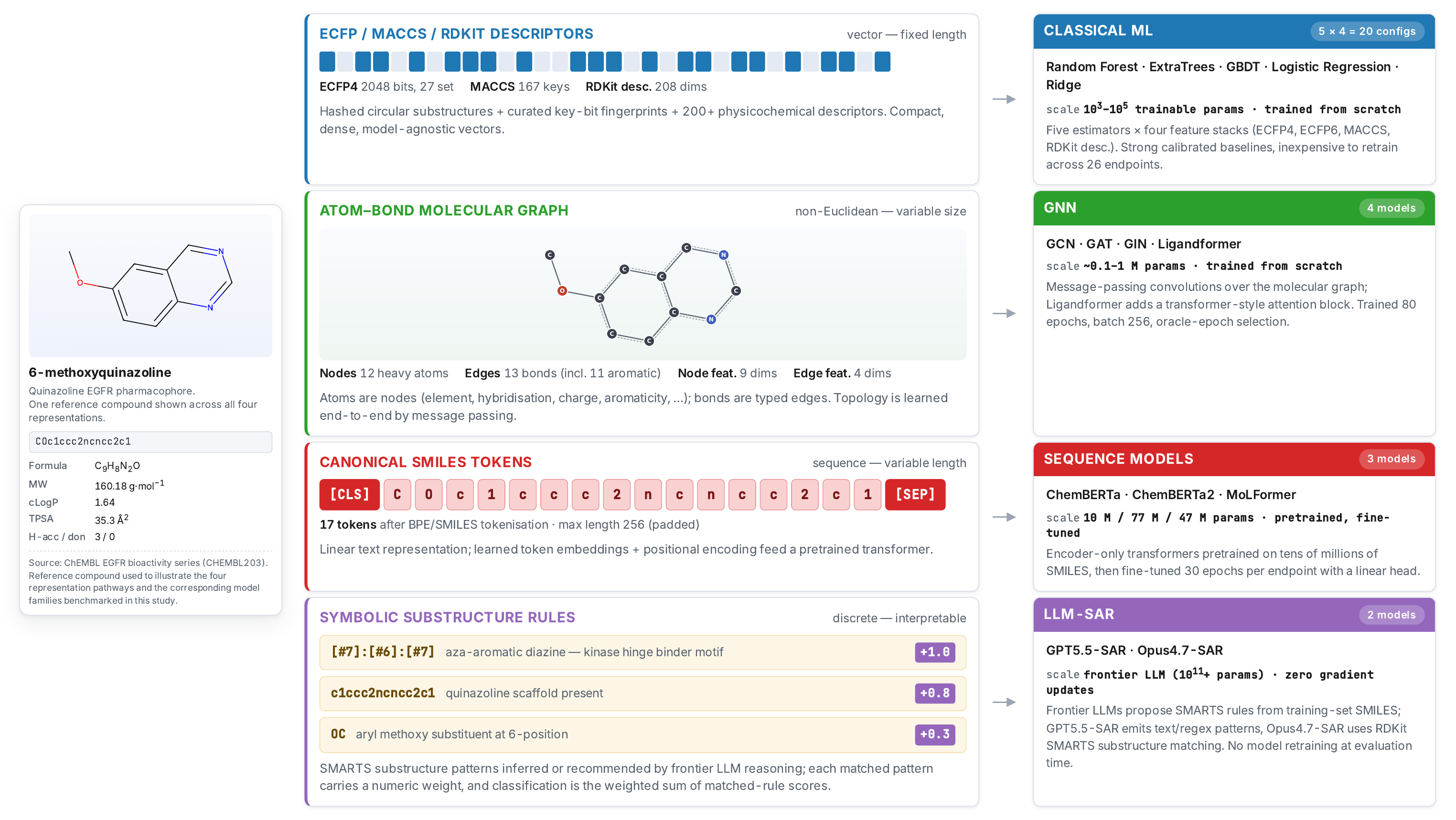}
\caption{Four model families and their corresponding molecular representation pathways compared in the benchmark, illustrated on a single representative compound (6-methoxyquinazoline; a quinazoline EGFR pharmacophore drawn from the CHEMBL203 series). The same molecule is encoded as (1) ECFP4 / MACCS / RDKit descriptor vectors feeding the classical ML family (RF / ExtraTrees / GBDT / LR / Ridge); (2) an atom bond molecular graph for the GNN family (GCN / GAT / GIN / Ligandformer); (3) tokenised canonical SMILES for the pretrained sequence family (ChemBERTa / ChemBERTa2 / MoLFormer); and (4) a list of matched symbolic substructure rules with additive contributions for the LLM-SAR family (GPT5.5-SAR / Opus4.7-SAR).}
\label{fig:representation-pathways}
\end{figure}

We compare four model families that span distinct molecular representations and inductive biases (Fig.~\ref{fig:representation-pathways}). Their differences are not reducible to model size. Fingerprints, molecular graphs, SMILES tokens and explicit SAR rules encode different assumptions about which chemical patterns should generalize across endpoints and split protocols. This distinction is central to interpreting the benchmark, so the implementation settings are kept constant within each family while allowing them to differ across families when their representations and optimization requirements differ.

\subsection{Classical Fingerprint and Descriptor Machine Learning}

The fingerprint and descriptor family contains classical QSAR learners trained on fixed molecular representations. Random forest (RF) uses bagged decision trees with randomized feature selection \citep{breiman2001rf}. ExtraTrees increases this randomization by sampling split thresholds more aggressively \citep{geurts2006extratrees}. Gradient boosting decision trees (GBDT) build additive tree ensembles by sequentially fitting residual errors \citep{friedman2001gbm}. Logistic regression and ridge regression provide linear baselines. Support vector machines and XGBoost are related classical baselines that are discussed as context but are not included in these result tables \citep{cortes1995svm,chen2016xgboost}.

RF and ExtraTrees are trained as 300 tree ensembles on ECFP4, ECFP6, MACCS keys and RDKit descriptor representations. Classification models use balanced class weights where supported; GBDT uses balanced sample weights. ECFP fingerprints encode atom centred neighbourhoods and map local substructures into sparse fixed length vectors, making them useful for similarity search, visualization and QSAR \citep{rogers2010ecfp,riniker2013similaritymaps}. This family is small in parameter count, but it is not a weak baseline: chemical featurization can give RF and ExtraTrees a strong advantage in low data, imbalanced or local SAR regimes.

\subsection{Graph Neural Networks}

The graph family represents each molecule as a graph with atoms as nodes and bonds as edges. GCN, GAT and GIN are trained as task specific molecular graph predictors with roughly $10^5$ to $10^6$ learned parameters. Ligandformer is included as a self attention graph model with approximately 1.4M parameters. To keep the GNN comparison reproducible, all graph models are trained under fixed optimization settings across endpoints and split protocols: up to 80 epochs, batch size 256, Adam optimisation with learning rate $10^{-3}$, and fold specific positive class weights clipped at 50 for imbalanced classification tasks.

Message passing neural networks provide the general molecular GNN abstraction: node representations are updated by aggregating information from neighbouring atoms and then pooled to form a graph level representation \citep{gilmer2017mpnn,fang2023gnnreview}. GCN uses normalized neighbourhood convolution \citep{gcn2017}; GAT learns attention weights over neighbouring nodes \citep{gat2018}; and GIN uses an expressive aggregation function motivated by the Weisfeiler Lehman graph isomorphism test \citep{gin2019}. Ligandformer adds a self attention graph backbone to this family \citep{ligandformer2022}. These models learn task specific molecular representations and may capture endpoint specific graph patterns that fixed fingerprints miss.

\FloatBarrier

\subsection{Pretrained Molecular Sequence Models}

The pretrained molecular sequence family treats SMILES as a chemical language. ChemBERTa, ChemBERTa2 and MoLFormer are fine tuned per endpoint from pretrained checkpoints with classification or regression heads. ChemBERTa and ChemBERTa2 use 77M and 10M parameter checkpoints, respectively, while MoLFormer has approximately 47M parameters. Fine tuning uses maximum sequence length 256, the AdamW optimiser with learning rate $2\times10^{-5}$ and per fold positive class weights clipped at 50; ChemBERTa and ChemBERTa2 use batch size 64, and MoLFormer uses batch size 8. All three sequence models are fine tuned for up to 30 epochs per fold under fixed optimization settings shared across endpoints and split protocols.

ChemBERTa adapts transformer pretraining to molecular property prediction and was pretrained on large PubChem SMILES corpora \citep{chemberta2020}. ChemBERTa2 extends this line towards chemical foundation models \citep{chemberta22022}. MoLFormer uses large scale molecular language pretraining to capture structural regularities from SMILES sequences \citep{molformer2022}. Their advantage is broad unsupervised chemical exposure. Their risk is representational mismatch: SMILES pretraining may not align with a narrow assay endpoint or a chemically shifted test fold.

\subsection{LLM-SAR Rule Baselines}

The LLM reasoning family tests whether frontier model chemical reasoning can be converted into deterministic SAR predictors. These baselines are not trained predictors in the same sense as the classical ML, GNN and sequence models. They are best interpreted against the broader question of what LLMs can and cannot do in chemistry \citep{guo2023llmchem}. GPT5.5-SAR and Opus4.7-SAR denote two independent SAR rule generation workflows that originate from two different frontier large language models. The rules and per rule scoring decisions were obtained \emph{once}, by asking each frontier model to reason over molecular structures, endpoint biology and structure activity relationships, and to emit (a) a fixed rule library that can be applied deterministically to a SMILES string and (b) a scoring rubric that maps rule firing patterns to a held out prediction.

GPT5.5-SAR refers to the regex on SMILES rule library derived from GPT5.5; Opus4.7-SAR refers to the RDKit SMARTS rule library derived from Claude Opus 4.7, with explicit endpoint priors. After SAR rule generation, both rule libraries are frozen, version controlled and applied locally; no held out molecule is ever sent to a live LLM API at evaluation time. This design preserves the large model reasoning component while making the held out prediction stage reproducible.

Each LLM-SAR family is evaluated in two modes. The first uses the expert SAR rules generated by the LLM without fold-derived knowledge: functional groups, aromatic and heteroaromatic patterns, polarity, lipophilicity, hydrogen bond capacity, known toxicophores and metabolism sensitive motifs. The second mode adds SAR knowledge derived from the training fold: for each split and each fold, training structures and labels are summarized into endpoint specific SAR or structure property relationship (SPR) knowledge, those learned rules are combined with the medicinal chemistry priors generated by the LLM, and the joint rule set is applied to score the held out fold. This design tests whether LLM reasoning, optionally augmented with rules derived from the training set, can compete with fitted statistical models. The results indicate that learned SAR knowledge often helps both LLM-SAR families, but does not replace supervised predictors when labelled training data are available; this is developed quantitatively in the LLM-SAR Results section below.

\section{Evaluation Protocol}
\label{sec:eval-protocol}

\begin{figure}[!htbp]
\centering
\includegraphics[width=0.96\linewidth]{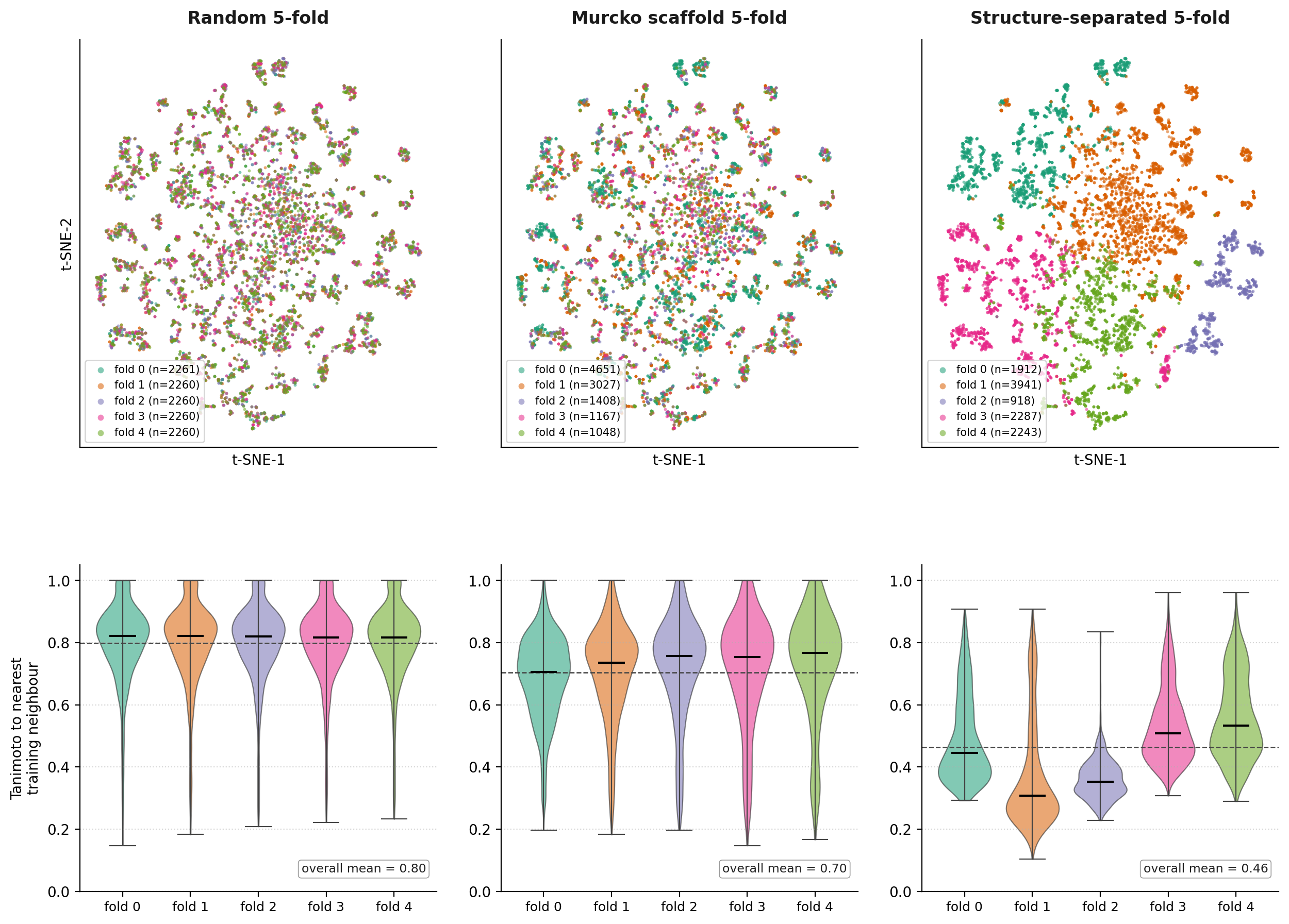}
\caption{Representative comparison of the three split protocols using the CHEMBL203 EGFR kinase inhibition panel as an example (11{,}301 unique compounds after canonical SMILES standardisation and conflict resolution). Top: ECFP4 vectors projected to 2D by t-SNE (perplexity 30, initialised from PCA), coloured by fold under random, Murcko scaffold and structure separated 5-fold CV. Bottom: Tanimoto similarity between each held out compound and its nearest training fold neighbour, shown separately for each held out fold. Random and Murcko folds overlap geometrically in ECFP space, but their overall mean nearest neighbour Tanimoto similarities differ (0.80 vs 0.70); structure separated folds are both geometrically and chemically further from train (overall mean 0.46), making the structure split a stronger test of out of distribution generalisation.}
\label{fig:split-panels}
\end{figure}

The benchmark is organized around three 5-fold cross validation protocols. They are ordered from the least chemically separated to the most chemically separated setting, and therefore probe increasing levels of generalization difficulty (Fig.~\ref{fig:split-panels}).

\begin{itemize}[leftmargin=*, itemsep=0.35em]
\item \textbf{Random 5-fold CV.} Molecules are assigned to folds without enforcing structural separation. Close analogues can therefore appear in both train and test folds. This is usually the easiest setting because models can interpolate within familiar local chemical neighbourhoods. It approximates retrospective evaluation inside a chemically well sampled project library.

\item \textbf{Murcko scaffold 5-fold CV.} Molecules are grouped by Bemis Murcko scaffold before fold assignment \citep{bemis1996frameworks}. Compounds with the same core scaffold are kept in the same fold, so the held out molecules are more scaffold distinct from the training set. This is typically harder than random CV because it tests scaffold transfer rather than analogue interpolation. It approximates hit to lead expansion, where models must generalize from explored scaffolds to related but unseen cores.

\item \textbf{Structure separated 5-fold CV.} Molecules are standardized and deduplicated by canonical SMILES; classification duplicates with conflicting labels are removed. ECFP4 fingerprints are first compressed with TruncatedSVD and then clustered with MiniBatchKMeans. Fold assignment is performed at the cluster level rather than the molecule level, so molecules occupying the same local region of chemical space are kept in the same fold. This reduces near neighbour leakage between training and held out folds and makes the split a stronger test of chemical extrapolation. It is the most stringent setting in the benchmark because held out folds occupy separated regions of chemical space, a scenario closer to library expansion or hit identification on new chemical matter.
\end{itemize}

All three protocols use five folds and no internal validation subset. For each endpoint and split, models are trained on four folds and evaluated on the held out fold. Classical ML uses ordinary fold means, and LLM-SAR is deterministic after rule generation. Epoch dependent GNN and sequence models are summarized using two complementary readouts. The primary optimal held-out readout reports the best held-out fold metric within fixed training settings and is used for the main benchmark tables. A fixed final-window readout, defined as the mean of the final three epochs from the same training logs, is reported as a supplementary protocol sensitivity analysis to test whether strict winner assignments depend on model selection. Every endpoint and split entry used for model family comparison includes classical ML, GNN, pretrained molecular sequence and LLM-SAR results.

Each split is constructed independently for each endpoint. No molecule from the held out fold is used to construct training fold SAR knowledge, tune rule thresholds or fit model parameters. This split specific treatment is important for comparing LLM-SAR with learned knowledge: random split knowledge, Murcko split knowledge and structure split knowledge are never pooled.

Classification tasks are primarily compared by PR-AUC, with ROC-AUC, class specific precision and recall, accuracy and top K enrichment used as supporting metrics. PR-AUC is emphasized for imbalanced toxicity and activity assays because its random baseline is the held out positive rate. Regression tasks are compared by MAE and Pearson correlation, with Spearman correlation and RMSE retained in the metric tables. Calibration summaries are computed for classification tasks to separate ranking performance from probability reliability.

Statistical robustness is tested with paired bootstrap resampling. For each pair of models $(A, B)$ on the same task, we draw 1{,}000 resamples of the held out predictions, compute $\Delta(\mathrm{metric}) = \mathrm{metric}(A) - \mathrm{metric}(B)$ on each resample, and call the comparison significant if the 95\% confidence interval of $\Delta$ excludes zero. Classification metrics use stratified resampling where possible. Because many comparisons are evaluated, $p$ values are adjusted using the Benjamini and Hochberg false discovery rate correction at $q = 0.05$, which controls the expected fraction of false positives among the comparisons declared significant. Comparisons are organised in two levels that match the manuscript's claim hierarchy. The \textbf{family level} compares the best model within each family on the same task across classical ML, GNN, sequence and LLM-SAR, and bears on the family level conclusions of Section~\ref{sec:results-family-wins}. The \textbf{model level} compares the three pairwise contrasts among the top ranked, second ranked and third ranked models where available, and bears on model level conclusions for individual tasks. The two levels are corrected separately, so within each level and metric a fixed $q = 0.05$ BH FDR threshold defines significance. This analysis distinguishes stable family level claims from close model level rankings.

\subsection{Implementation Caveats}

Four implementation choices affect interpretation. First, GBDT does not support a native \texttt{class\_weight} argument in scikit-learn, so balanced \texttt{sample\_weight} is supplied during fitting; RF, ExtraTrees and LR use balanced class weights directly. Second, GNN and sequence classification losses use training fold positive class weights clipped at 50 to stabilize optimization on highly imbalanced folds. Third, fold quality flags identify small or chemically uneven held out folds across multiple endpoints; representative cases include Caco2\_Wang fold 0 being marked \texttt{small\_test} and PXR/NR1I2 structure separated folds with extreme class imbalance. Fold quality flags for each endpoint are released as machine readable supplementary data so that downstream readers can inspect any individual fold. Fourth, the two SAR baselines differ in SMILES handling: GPT5.5-SAR is text rule oriented, whereas Opus4.7-SAR uses RDKit SMARTS patterns. These differences are methodologically meaningful and do not constitute a direct comparison of commercial LLM APIs. The benchmark therefore follows QSAR best practice expectations by reporting fold construction, model settings and metric definitions explicitly \citep{tropsha2010qsar}.

\FloatBarrier

\section{Results}

\subsection{Model Family Win Analysis Across Endpoint Classes}
\label{sec:results-family-wins}

\begin{figure}[!htbp]
\centering
\includegraphics[width=0.86\linewidth,height=0.42\textheight,keepaspectratio]{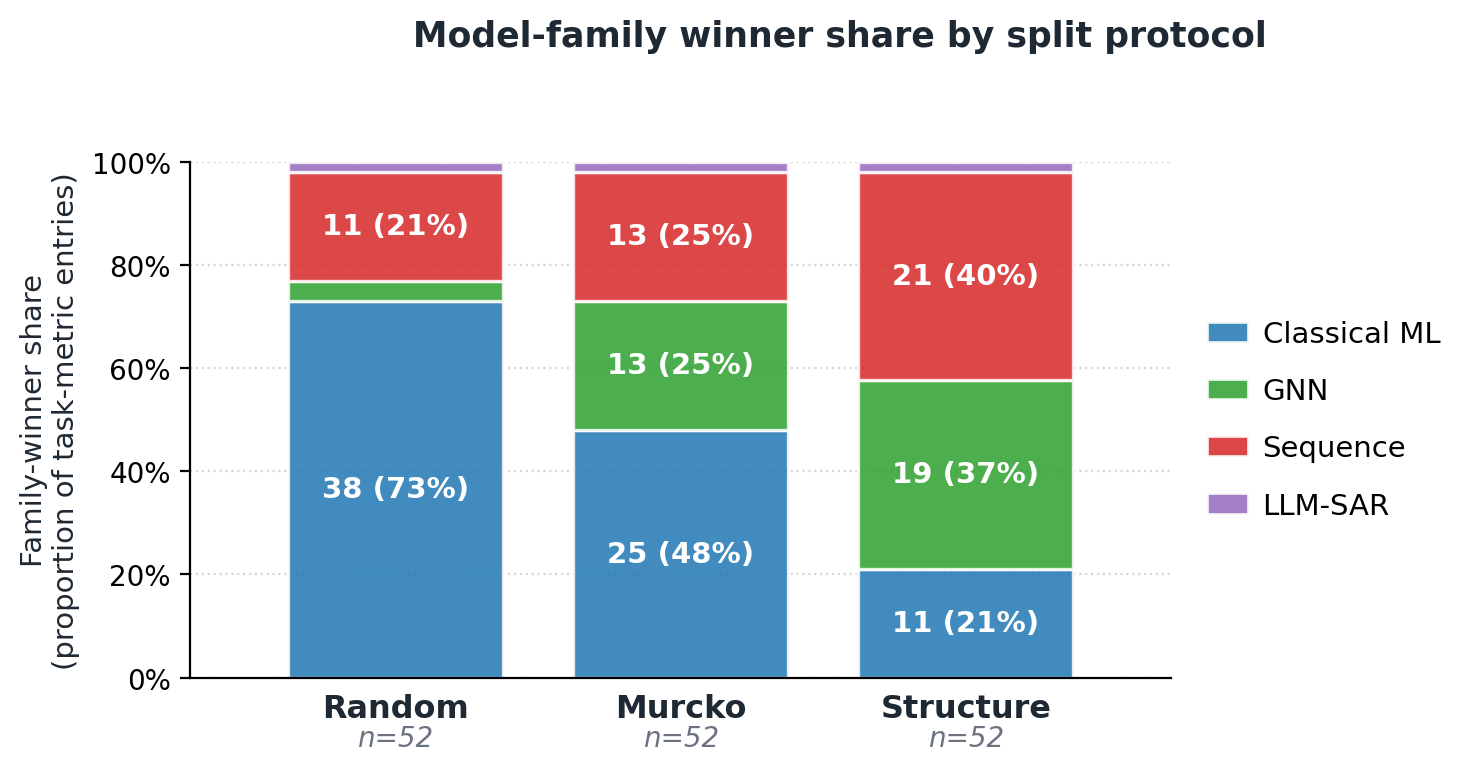}
\caption{Model family winner share by split protocol under the primary optimal held-out readout. Each bar is the proportion of task and metric entries within one split that are won by each family, pooled across all 26 endpoints. Wins are decided jointly across PR-AUC and ROC-AUC for classification endpoints and MAE and Pearson for regression endpoints, so each task and metric entry contributes one strict first-place assignment. The number under each bar gives the total task and metric entries pooled for that split ($n=52$ per split: 26 endpoints $\times$ 2 metrics). Counts are based on fold aligned classical ML, GNN, sequence model and LLM-SAR results, with each family contributing its best individual model or LLM-SAR rule variant per task and metric entry. The corresponding split by endpoint class counts are reported in Table~\ref{tab:family-winner-summary}.}
\label{fig:family-winners-stacked}
\end{figure}

The unified endpoint panel gives a consistent view across molecular property, toxicity, liability and activity prediction. Across 156 fold mean task and metric comparisons under the primary optimal held-out readout, classical ML wins 74 comparisons (47.4\%), pretrained molecular sequence models win 45 (28.8\%), GNNs win 34 (21.8\%), and LLM-SAR wins 3 (1.9\%); the four LLM-SAR variants enter the comparison through a single family best score per task and metric entry (Table~\ref{tab:family-winner-summary}). The split resolved summary in Fig.~\ref{fig:family-winners-stacked} shows that this aggregate pattern changes substantially across CV protocols. Under random CV, classical ML wins 73.1\% of task and metric comparisons, consistent with the easier interpolation setting. Under Murcko scaffold CV, classical ML still contributes the largest share (48.1\%), but GNN and sequence models together account for 50.0\%. Under structure separated CV, sequence models win 40.4\% of comparisons, GNNs win 36.5\%, and classical ML accounts for 21.2\%. These strict first-place counts summarize which family ranks first under the primary optimal held-out readout, but each task and metric entry contributes one winner regardless of the numerical margin. A fixed final-window sensitivity analysis gives a more conservative view of GNN and sequence model first-place assignments: their winner shares decrease when epoch selection is replaced by a fixed final-window readout, whereas classical ML is unchanged by this readout choice (Supplementary Note~2b). Thus, strict winner counts provide a useful family-level summary, but the exact distribution of first-place assignments depends partly on readout choice and should not be interpreted as a definitive ranking. LLM-SAR wins remain rare and are confined to toxicity related comparisons (Table~\ref{tab:family-winner-summary}). Because PR-AUC depends on positive class prevalence, PR-AUC based wins are interpreted within each endpoint rather than as absolute magnitudes across endpoints; this is particularly relevant for high prevalence endpoints such as EGFR, DRD2 and PXR/NR1I2. Public anti-TB H37Rv and antimalaria Pf. 3D7/Dd2 phenotypic activity contribute to the bioactivity related class; restricted institutional infectious disease datasets are reported separately as supplementary restricted data analyses (Supplementary Note~5) and are excluded from the primary family win denominator.

To make the family level win analysis transparent at the individual model level, Tables~\ref{tab:table-structure-adme-prauc-rocauc-model-matrix}--\ref{tab:table-structure-bioactivity-prauc-rocauc-model-matrix} report model by endpoint matrices under structure separated CV for ADME related, toxicity related and bioactivity related endpoints. Classification and regression endpoints are separated. Classification tables report PR-AUC / ROC-AUC within each endpoint entry. The ADME related regression table reports MAE / Pearson within each endpoint entry. Because the toxicity related classification panel is larger and mechanistically heterogeneous, it is divided into general toxicity and safety liability endpoints, Tox21 nuclear receptor assays and Tox21 stress response assays. Each table caption states the metric pair used. Rankings are computed independently for each metric within each endpoint: the first ranked value is bold and underlined, and the second and third ranked values are underlined. Complete random split and Murcko scaffold split matrices are provided in Supplementary Note~2.

This result should not be read as a claim that one smaller model family is always best. Rather, the results indicate that endpoint level performance is not monotonic with model scale. Classical ML models, especially tree ensembles using ECFP fingerprints or RDKit descriptors, provide the most stable leading family across the benchmark. GNNs and pretrained molecular sequence models remain competitive in selected harder split protocols, but their gains are endpoint dependent and partly sensitive to readout choice. The best performing family is therefore an empirical property of the endpoint, split protocol and molecular representation, rather than a direct consequence of parameter count.
\FloatBarrier

\subsection{Split realism and the difficulty axis}

\begin{figure}[!htbp]
\centering
\includegraphics[width=\linewidth]{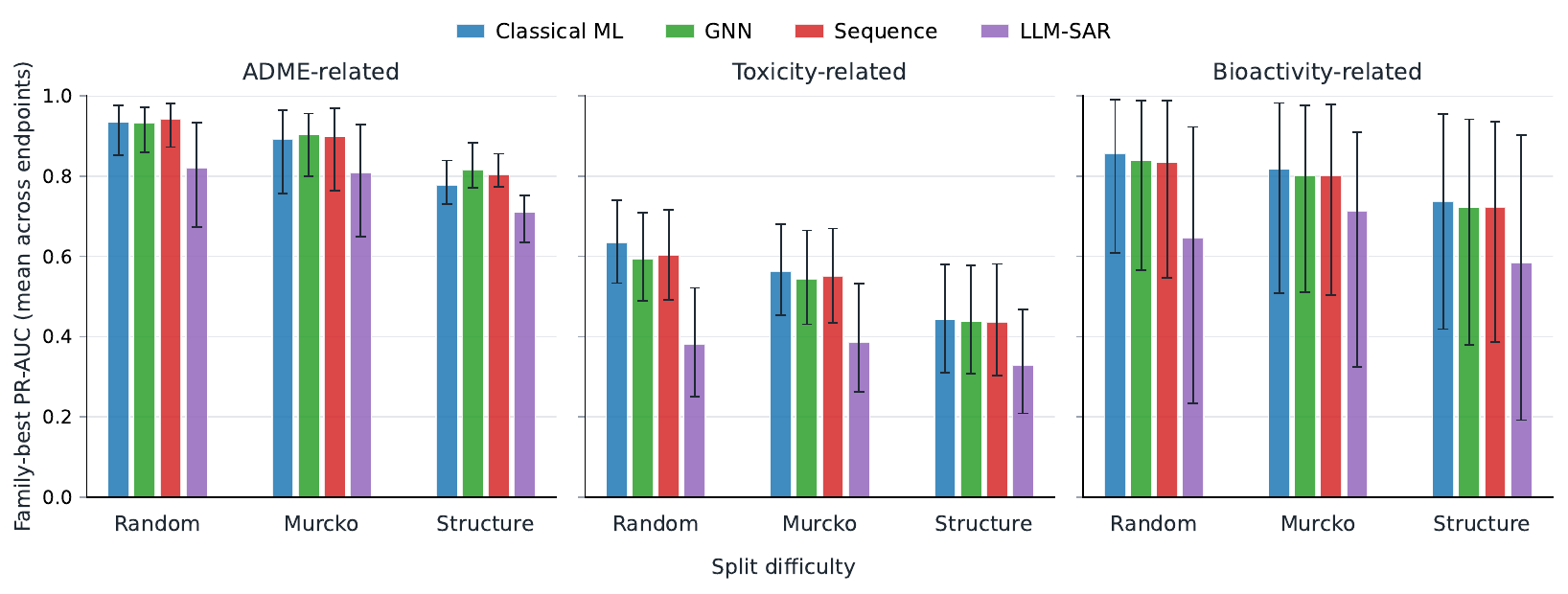}
\caption{Mean family best PR-AUC as a function of split difficulty, panelled by endpoint class. Bars report means across endpoint level family best values; error bars show bootstrap 95\% confidence intervals. The split axis is ordered by realism for drug discovery deployment: random (easiest, late stage retrospective on a closed library) $\rightarrow$ Murcko scaffold (intermediate, scaffold expansion in hit to lead) $\rightarrow$ structure separated (hardest, library expansion or hit identification on novel chemotypes).}
\label{fig:split-realism}
\end{figure}

Using the difficulty ordering defined in the Evaluation Protocol, we next ask how each family's predictive performance changes as the chemical separation between training and test folds becomes more demanding. We treat the three split protocols as both a difficulty axis and a representation stress test: how much does each family's PR-AUC drop when train and test chemistry are forced apart, and which family loses the most ground in absolute terms? Figure~\ref{fig:split-realism} reports mean family best PR-AUC with bootstrap 95\% confidence intervals at each split level, and Table~\ref{tab:split-realism-drops} provides the corresponding mean drop from random to structure separated CV.

Across endpoint classes, increasing split realism generally lowers predictive performance. Random splits are the easiest setting because training and test folds can contain close analogues from the same local SAR neighbourhoods. Murcko scaffold splits remove part of this overlap by holding out scaffold frameworks, and structure separated splits create the strongest chemical shift by holding out more distant chemotypes. As structural separation increases, classical ML remains the largest winner family in aggregate (Table~\ref{tab:family-winner-summary}), but its random split advantage narrows. From random to structure separated CV, mean family best PR-AUC for classical ML falls by 0.156 on ADME related, 0.193 on toxicity related and 0.119 on bioactivity related endpoints. GNNs fall by 0.117, 0.157 and 0.117, respectively, and pretrained sequence models fall by 0.139, 0.167 and 0.112. These drops indicate that all supervised families are affected by chemical separation. Under the primary optimal held-out readout, GNN and sequence models are competitive in selected harder split protocols, but the fixed final-window sensitivity analysis indicates that some strict first-place assignments in these settings are readout-sensitive. The robust conclusion is therefore not that GNN and sequence models are generally superior to classical ML under chemical separation, but that split difficulty lowers performance across all supervised families and makes many family differences numerically close. LLM-SAR remains the weakest family in absolute predictive performance, but its scores change less across the split axis, consistent with rule based structural alerts being less dependent on close analogue overlap.

Random 5-fold CV strongly favours classical ML models based on fingerprints and descriptors. Classical ML accounts for 38 of 52 task and metric wins under random splitting and reaches mean family best PR-AUC of 0.934 / 0.635 / 0.856 on ADME related, toxicity related and bioactivity related endpoints. ECFP, MACCS and RDKit descriptors expose local substructures, physicochemical patterns and analogue series membership directly to RF, ExtraTrees and GBDT. When training and test folds contain close analogues, these fixed molecular representations are often sufficient: tree ensembles can interpolate within familiar local SAR neighbourhoods rather than learning a new representation from the data.

ADME related endpoints show the clearest shift away from the random split pattern under the primary optimal held-out readout. In the PR-AUC view, the three supervised families are already close under random CV, with mean family best values of 0.934 for classical ML, 0.933 for GNN and 0.943 for sequence models. Under structure separated CV, GNN reaches the highest mean family best PR-AUC (0.816), followed by sequence models (0.804) and classical ML (0.779). The corresponding mean drops from random to structure separated CV are 0.117 for GNN, 0.139 for sequence models and 0.156 for classical ML. This pattern suggests that molecular graph topology and pretrained SMILES representations are less penalised than fixed fingerprint and descriptor models on these ADME related classification endpoints in the primary optimal held-out readout, although the confidence intervals remain broad because this class contains a small number of PR-AUC endpoints and the fixed final-window sensitivity analysis treats strict winner assignments as readout-sensitive. Chemically, this is plausible because ADME labels often depend on combinations of physicochemical and structural factors, including permeability, ionisation, metabolic liability and receptor mediated effects, rather than on a single local structural alert.

Toxicity related endpoints show a more nuanced balance. Classical ML remains the leading family at every split level (mean PR-AUC 0.635, 0.563, 0.442) and incurs the largest absolute drop in the class (0.193), but the performance gap to both GNN and sequence families narrows as the split becomes harder. Sequence models are 0.032, 0.012 and 0.005 PR-AUC below classical ML under random, Murcko scaffold and structure separated CV, respectively; the corresponding gaps for GNNs are 0.041, 0.019 and 0.004. Many toxicity tasks are imbalanced and depend on local alerts, electrophilic motifs, polyaromaticity, lipophilicity or receptor specific substructures that ECFP/RDKit features expose. These alerts persist across scaffolds, which is why classical ML retains the lead even as its absolute score falls. GNN and sequence models still win selected toxicity entries, especially on pathway specific Tox21 assays where graph topology or pretrained SMILES patterns align with the underlying receptor or stress response mechanism.

Bioactivity related endpoints illustrate the endpoint dependence of this analysis. The class combines a target based EGFR kinase inhibition assay with phenotypic anti-TB H37Rv and antimalaria Pf. 3D7/Dd2 cellular activity endpoints, so the apparent split effect is shaped by both target specific SAR and phenotypic activity mechanisms. In the mean PR-AUC view, classical ML, GNN and sequence models are close under structure separated CV (0.737, 0.722 and 0.723 respectively), indicating no large family level separation after chemical shift. The mean drops from random to structure separated CV are also similar across the three supervised families (0.119, 0.117 and 0.112). Thus, bioactivity does not support a single family winner under the split realism analysis: classical ML retains a small mean advantage, whereas GNN and sequence models remain close. The wide confidence intervals in Fig.~\ref{fig:split-realism} reflect the small number of bioactivity endpoints and their heterogeneous mechanisms.

LLM-SAR shows the most distinctive split realism signature. It has the lowest winner count across the split axis, with three task and metric wins in total, but it also has the smallest split induced drop. On toxicity related endpoints, LLM-SAR mean family best PR-AUC is 0.381 under random, 0.387 under Murcko and 0.329 under structure separated CV, for a drop of 0.052 from random to structure separated CV. A small reversal at the Murcko level is consistent with chemotype specific alert frequency. On ADME related and bioactivity related endpoints the mean drops are 0.111 and 0.063, both smaller than the corresponding supervised family drops. The mechanism is methodological: LLM-SAR rules are derived from chemotype agnostic medicinal chemistry priors, including toxicophores, structural alerts and electrophilic motifs. Removing chemical similarity between train and test folds removes classical ML's analogue interpolation advantage but does not remove these alerts. Within the toxicity class, the gap between LLM-SAR and classical ML therefore shrinks from 0.254 under random CV to 0.114 under structure separated CV. The implication is that LLM-SAR's value is best read as robustness under chemical distribution shift, not as headline accuracy: it complements supervised predictors in the deployment regimes where they are most exposed to shift. Overall, split realism does not support a simple rule that larger models win. The primary optimal held-out readout highlights selected settings in which GNN and sequence models are competitive, whereas the fixed final-window readout identifies classical ML as the most stable leading family. LLM-SAR is stable across the split axis but remains lowest in absolute predictive performance.

\FloatBarrier

\subsection{Top K Enrichment and Virtual Screening Utility}

Beyond ranking summaries, virtual screening triage requires that a small top $K$ fraction of a screening library captures a large fraction of true actives. We therefore compute EF@1\%, EF@5\% and EF@10\% for every family best model on every split and task entry. EF@$K$\% measures how many times more active molecules a model promotes into the top $K$\% of its ranked list compared with random screening. EF@1\% represents a stringent early recognition setting; EF@5\% and EF@10\% represent progressively broader triage thresholds and are retained as machine readable supporting metrics in Table~\ref{tab:topk-best-ef1}. The figure focuses on EF@1\% to keep early enrichment, which is the most consequential triage decision, directly comparable across the three split protocols. Top 50 and top 100 precision are also retained as supporting metrics.

\begin{figure}[!htbp]
\centering
\includegraphics[width=\linewidth]{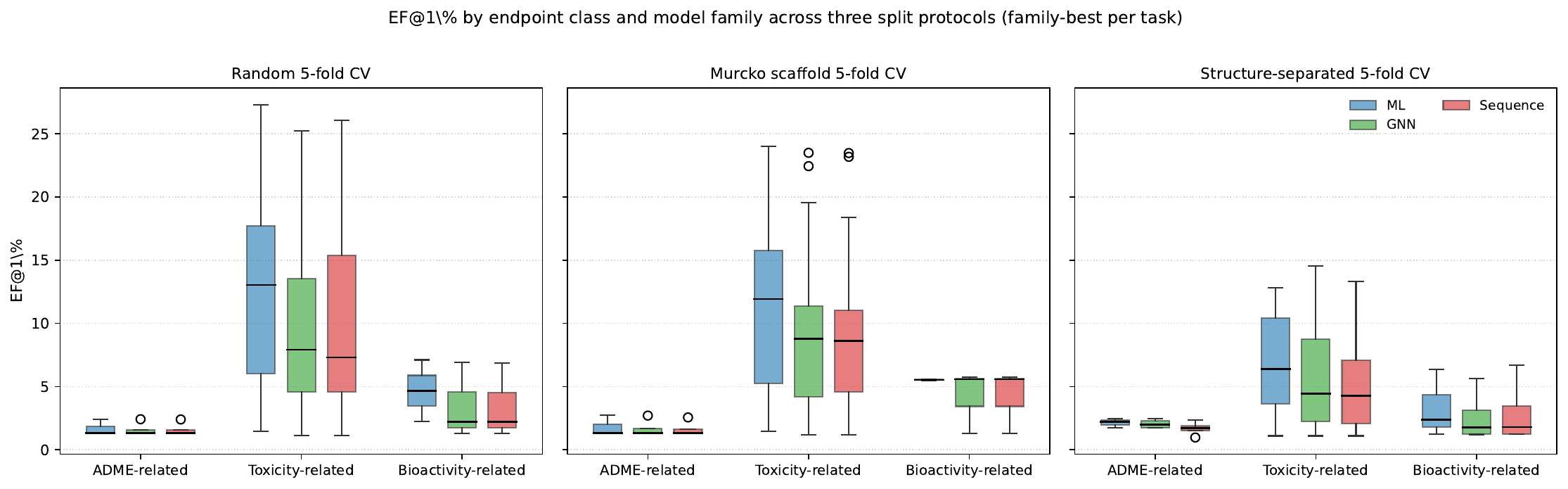}
\caption{EF@1\% by endpoint class and model family across the three split protocols. Each panel corresponds to one split (random, Murcko scaffold and structure separated 5-fold CV). Within each panel, boxes show the spread of family best EF@1\% across the tasks in each endpoint class. EF@1\% reports the fold mean enrichment of actives in the top 1\% of the ranked list compared with random selection. The family colour key is shared across panels.}
\label{fig:topk-ef}
\end{figure}

Fig.~\ref{fig:topk-ef} summarises family best EF@1\% across the three split protocols, with one panel per split; Table~\ref{tab:topk-best-ef1} reports the corresponding family level medians and best EF@1\% leaders pooled across splits. The split panels reveal that enrichment is strongly modulated by training and test chemical separation. Under random 5-fold CV, classical ML leads on every endpoint class, with toxicity related median EF@1\% of 13.0 (classical ML), 7.9 (GNN) and 7.3 (sequence), and a single endpoint best of EF@1\% 27.3 on tox21\_NR-AR-LBD (GBDT on ECFP6). Under Murcko scaffold CV the toxicity advantage compresses to 11.9 (classical ML), 8.8 (GNN) and 8.6 (sequence), while public anti-TB H37Rv and antimalaria Pf. 3D7/Dd2 both retain early enrichment after scaffold separation. Under structure separated CV, toxicity related median EF@1\% falls for every family (classical ML 6.4, GNN 4.5 and sequence 4.3), whereas public anti-TB H37Rv remains enriched by the best models from all three supervised families (sequence 6.69, classical ML 6.37 and GNN 5.61). Antimalaria Pf. 3D7/Dd2 shows more modest but still positive structure separated enrichment (classical ML 2.36, sequence 2.36 and GNN 2.32). ADME related enrichment values are smaller in absolute terms and show less consistent monotonic degradation across split protocols. Two general observations follow. First, enrichment is most sensitive to split protocol on toxicity related endpoints. Second, GNN, sequence and classical ML can all recover early active enrichment on public anti-infective cellular activity tasks, while classical ML retains a meaningful enrichment lead on toxicity related endpoints across all three splits. LLM-SAR is omitted from the panels because molecule level probability scores from deterministic rule application populate a coarse set of bins; the LLM-SAR section discusses this separately.

\FloatBarrier

\subsection{Training Data Regime Scaling}

\begin{figure}[!htbp]
\centering
\includegraphics[width=0.88\linewidth,height=0.46\textheight,keepaspectratio]{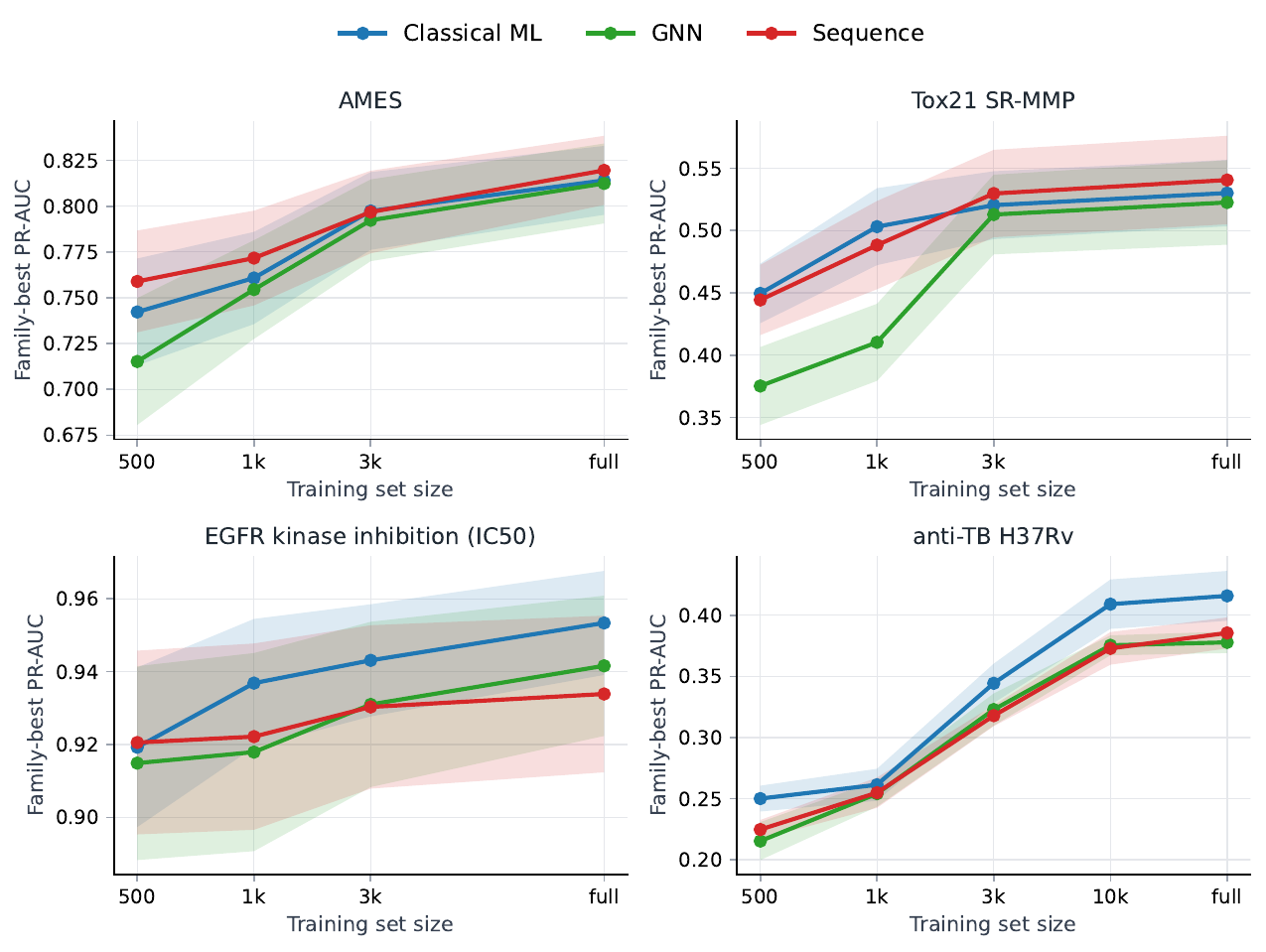}
\caption{Training data regime scaling for family best models. Lines show family best PR-AUC for the best available model in each family at each training size; shaded bands show fold level standard error. Across all four endpoint mechanisms (AMES mutagenicity, Tox21 SR-MMP, EGFR kinase inhibition and anti-TB H37Rv), classical ML remains broadly competitive across two orders of magnitude of training data: family best curves overlap within their fold level uncertainty bands at most sizes, the lead family flips between sizes on AMES and Tox21 SR-MMP, and classical ML holds a small but consistent lead on EGFR (size $\geq$ 1{,}000) and on anti-TB H37Rv at every size.}
\label{fig:data-regime}
\end{figure}

A focused data regime analysis evaluates whether family rankings change as labelled training size increases. The analysis includes rows at 500, 1{,}000, 3{,}000, 10{,}000 and full training sizes where each family and task combination is available; missing combinations are not extrapolated. Across the four representative tasks (AMES, Tox21 SR-MMP, EGFR kinase inhibition and anti-TB H37Rv), classical ML is broadly competitive at every training size, but the family ranking depends on the task. On AMES, the three families remain tightly clustered between PR-AUC 0.71 and 0.82 across all sizes, with classical ML, GNN and pretrained sequence models trading the lead; differences between any two families are smaller than the fold level standard deviation at every size, so the 95\% bands overlap throughout (Fig.~\ref{fig:data-regime}). On Tox21 SR-MMP, classical ML leads the smallest sizes (PR-AUC 0.450 at 500 compounds and 0.503 at 1{,}000 compounds, against 0.444 and 0.488 for the best sequence model), whereas pretrained sequence models take a small lead from 3{,}000 compounds onwards, with differences that remain within fold level uncertainty. On EGFR kinase inhibition, classical ML is the best family at every training size at and above 1{,}000 compounds, with RF(ECFP4) reaching PR-AUC 0.953 at the largest available size; at 500 compounds, the apparent MoLFormer advantage over RF(ECFP4) is 0.002 PR-AUC, within fold level noise. On anti-TB H37Rv, classical ML is the best family at every size from 500 to full, and its margin over the best GNN or sequence model is the most consistent of the four tasks, holding at approximately 0.020 to 0.030 PR-AUC across the size axis. Two general observations follow (Table~\ref{tab:data-regime-best}; Table~\ref{tab:data-regime-first-crossover}). First, when pretrained sequence models do lead, they tend to do so by margins smaller than the fold level noise, so the apparent winner can flip between adjacent training sizes within the same task. Second, classical ML remains a strong default across endpoint mechanisms and across two orders of magnitude of training data, with no evidence that simply adding data makes a larger or more pretrained model family the winner.

\FloatBarrier

\subsection{Prevalence Aware Leaders Across Selected Endpoints}

\begin{figure}[!htbp]
\centering
\includegraphics[width=\linewidth]{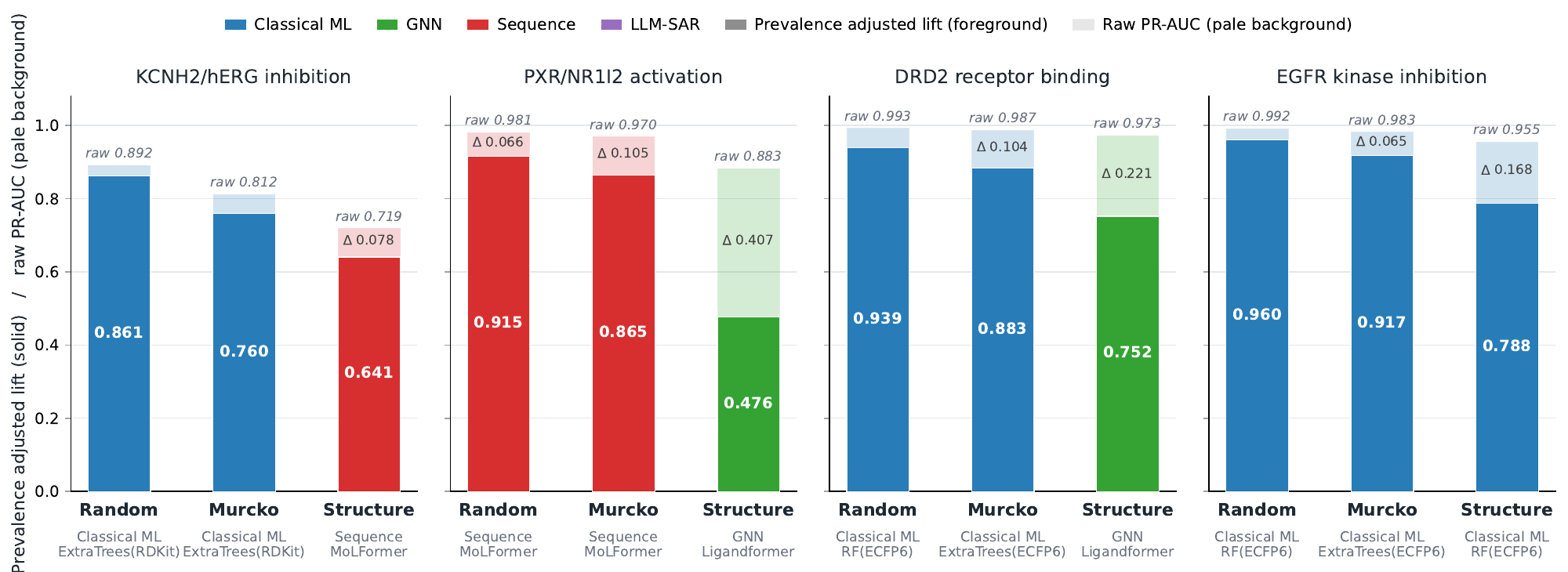}
\caption{Prevalence aware endpoint leaders. For each endpoint and split entry, two bars of identical width are overlaid for the family best model: a pale background bar (in the same family colour) reports the raw PR-AUC, and a solid foreground bar reports the prevalence adjusted lift $(\mathrm{PR\text{-}AUC} - p) / (1 - p)$, where $p$ is the held out positive rate. The visible shrinkage from background to foreground equals the prevalence inflation $\Delta = \mathrm{PR\text{-}AUC} - \mathrm{lift}$. Because the adjustment is a monotone affine transform within each endpoint and split entry, the family best identity is preserved; the adjusted view exposes how much of each raw PR-AUC is explained by the prevalence prior.}
\label{fig:prevalence-leaders}
\end{figure}

Several endpoints in the unified panel require prevalence aware interpretation (Table~\ref{tab:prevalence-leaders}; Fig.~\ref{fig:prevalence-leaders}). PXR/NR1I2 activation is ADME related, KCNH2/hERG inhibition and DRD2 receptor binding are toxicity related liabilities, and EGFR kinase inhibition is bioactivity related. Because these datasets are imbalanced (positive rates 77.7\%, 21.8\%, 88.9\% and 79.0\%), the leader table reports both PR-AUC and the prevalence adjusted lift $(\mathrm{PR\text{-}AUC} - p)/(1 - p)$, where \(p\) is the positive class prevalence; lift 1.00 corresponds to a perfect ranker and lift 0.00 to a model that does no better than the prevalence prior. Because prevalence adjustment is a monotone affine transform within each endpoint and split entry, the family best identity does not change between the raw and adjusted views; the adjusted view instead exposes how much of each raw PR-AUC is attributable to the prevalence prior. Across KCNH2/hERG inhibition, DRD2 receptor binding and EGFR kinase inhibition, classical ML models lead 7 of the 9 split and task entries, with RF(ECFP6) leading 3 entries (DRD2 random, EGFR random and EGFR structure separated) and ExtraTrees(ECFP6) and ExtraTrees(RDKit descriptors) leading 2 each. The remaining 2 entries are taken under structure separated CV by GNN and sequence families: MoLFormer leads KCNH2/hERG inhibition (PR-AUC 0.719, lift 0.64) and Ligandformer leads DRD2 receptor binding (PR-AUC 0.973, lift 0.75). Raw PR-AUC for the 9 entries ranges from 0.719 for KCNH2/hERG inhibition under structure separated CV to 0.993 for DRD2 receptor binding under random CV; the corresponding lifts range from 0.64 to 0.94, so even the lowest PR-AUC entry retains a sizeable lift over its 21.8\% prior. For PXR/NR1I2 activation, MoLFormer leads under random CV (PR-AUC 0.981, lift 0.92) and under Murcko scaffold CV (PR-AUC 0.970, lift 0.87), and Ligandformer leads under structure separated CV (PR-AUC 0.883, lift 0.48). These GNN and sequence leaders are reported under the primary optimal held-out readout; in the fixed final-window readout, the corresponding KCNH2/hERG, DRD2 and PXR/NR1I2 entries are led by classical ML models, so they should be read together with the readout sensitivity analysis in Supplementary Note~2b. The prevalence adjusted view therefore tempers raw PR-AUC values: high prevalence entries can still show substantial lift in random and Murcko folds, but the structure separated PXR/NR1I2 result, where the family best raw PR-AUC of 0.883 contracts to a lift of 0.48 once the 77.7\% prior is removed, illustrates why high prevalence endpoints require explicit prevalence baselines. A numerically high PR-AUC can correspond to modest lift when the positive class is already common.

\FloatBarrier

\subsection{Statistical robustness and calibration}

Paired bootstrap tests are used as a robustness check rather than as the main evidence for model choice. The paired bootstrap protocol and the BH FDR correction at $q = 0.05$ are defined in Section~\ref{sec:eval-protocol}. The analysis is organised in two levels. The \textbf{family level} compares the best model within each family on the same task across classical ML, GNN, pretrained sequence and LLM-SAR, and bears on the family level conclusions of Section~\ref{sec:results-family-wins}. The \textbf{model level} compares the three pairwise contrasts among the top ranked, second ranked and third ranked models where available, and bears on model level conclusions for individual tasks.

\begin{figure}[!htbp]
\centering
\includegraphics[width=0.82\linewidth,height=0.42\textheight,keepaspectratio]{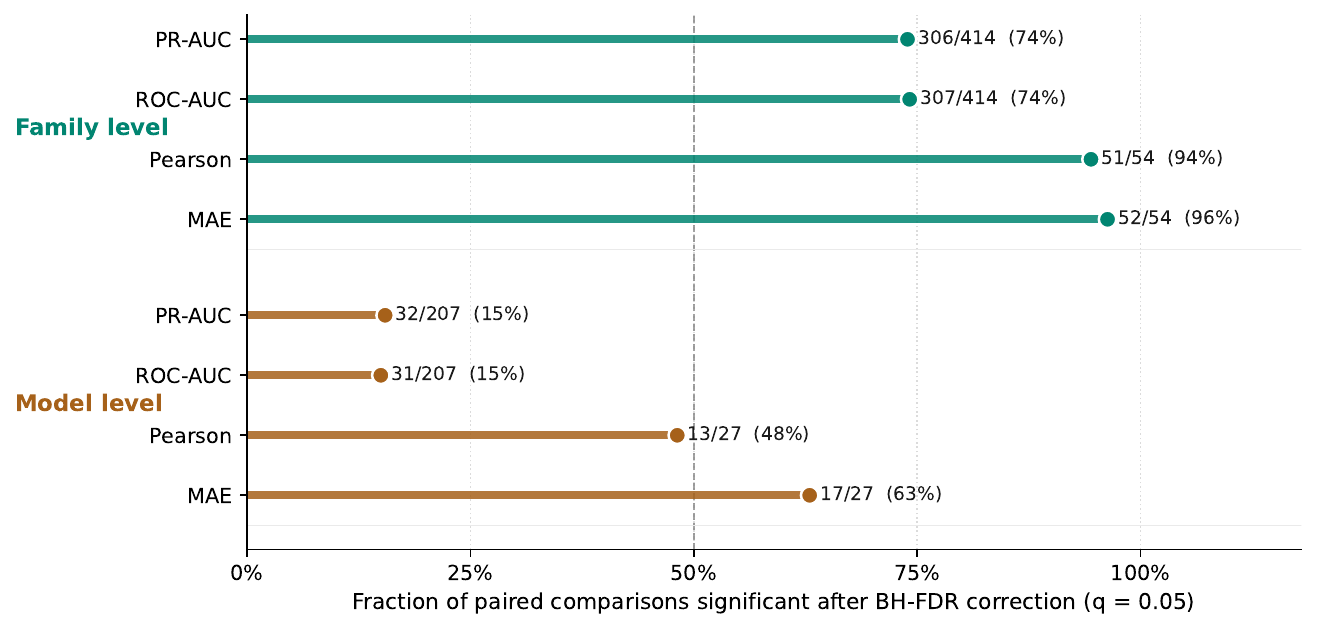}
\caption{Paired bootstrap significance summary across two test levels, the family level and the model level, and four metrics. Each row is one level and metric entry; the bar shows the fraction of paired comparisons whose $\Delta(\mathrm{metric})$ confidence interval excludes zero after Benjamini and Hochberg FDR correction at $q = 0.05$. The count to the right of each bar is significant tests over total tests. The dashed line at 50\% marks where a majority of tests is significant. The two levels are defined in Section~\ref{sec:eval-protocol}. Family level contrasts (green) are largely significant across all metrics, supporting the family level conclusions of Section~\ref{sec:results-family-wins}. Model level contrasts (orange) are mostly tied for classification metrics, so per task top three ranking claims should be treated as ranking guesses rather than statistically separable winners.}
\label{fig:paired-bootstrap}
\end{figure}

The results split sharply between the two levels (Table~\ref{tab:paired-bootstrap-headlines}; Fig.~\ref{fig:paired-bootstrap}). For PR-AUC, 306 of 414 family level contrasts (74\%) clear the BH FDR threshold, whereas 32 of 207 model level contrasts (15\%) do. ROC-AUC, Pearson and MAE show the same separation: family level fractions remain at 74 to 96\%, while model level fractions span 15\% to 63\% (PR-AUC 15\%, ROC-AUC 15\%, Pearson 48\%, MAE 63\%). The pattern means that family level comparisons are statistically supported, while per task model level claims about the top three ranking order should be read as ranking guesses rather than significant winners.

\begin{figure}[!htbp]
\centering
\includegraphics[width=\linewidth]{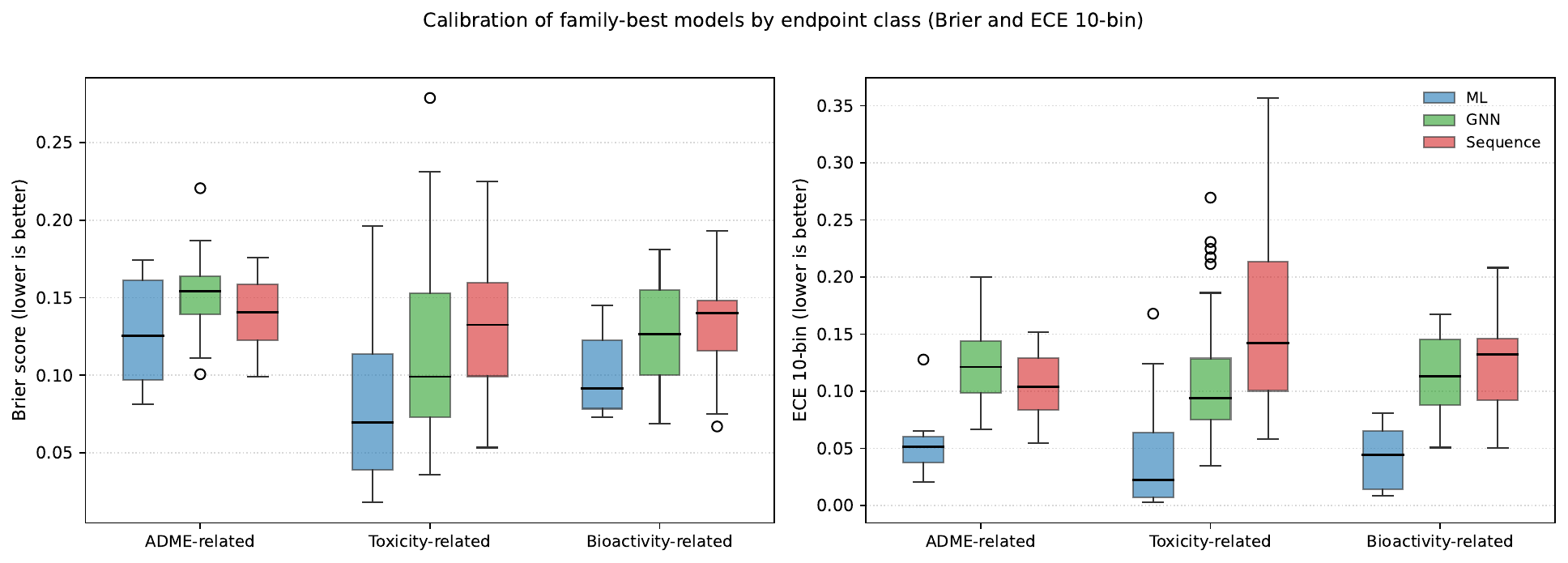}
\caption{Calibration of family best models by endpoint class. Boxes show Brier (left) and 10 bin ECE (right) for the best PR-AUC model in each family on each split and task entry. Lower is better in both panels.}
\label{fig:calibration}
\end{figure}

The same prediction pools that support the paired bootstrap tests also support a calibration analysis. We score Brier and the 10 bin Expected Calibration Error (ECE) for the best PR-AUC model in each family on each split and task entry. Classical ML is better calibrated than GNN and sequence models in all three endpoint classes (Fig.~\ref{fig:calibration}; Table~\ref{tab:calibration-best-by-class}). The difference is largest in toxicity related endpoints, where classical ML median Brier / ECE is approximately 0.070 / 0.022, compared with 0.099 / 0.094 for GNN and 0.132 / 0.142 for sequence models. The same ordering is observed for bioactivity related endpoints, with classical ML median Brier / ECE of 0.092 / 0.044, compared with 0.127 / 0.113 for GNN and 0.140 / 0.132 for sequence models. The results show that classical ML combines competitive ranking performance with better calibration across these endpoints, while high PR-AUC for GNN and sequence models does not necessarily imply reliable predicted probabilities. LLM-SAR is omitted from the panel because rule application returns a coarse score that does not behave as a calibrated probability; Brier and ECE are reported for classical ML/GNN/sequence models.

\FloatBarrier

\section{LLM-SAR: Useful Reasoning, Limited Prediction}
\label{sec:llm-sar}

The LLM-SAR baselines occupy a distinct role in this benchmark. They are not statistical models trained from labelled data; they are deterministic SAR rule libraries generated once from frontier large language models through extended chemical reasoning and then applied locally with no further LLM involvement at evaluation time. This subsection describes the applicable scope of LLM-SAR, the practical value of the generated SAR rule library, and how SAR knowledge derived from training folds changes the prediction quality.

\paragraph{Applicable scope.}
LLM-SAR is most informative when (i) the endpoint admits a relatively small library of well recognised medicinal chemistry alerts or SAR motifs, (ii) the dataset is small or imbalanced enough that a fitted statistical model is operating near its noise floor, or (iii) explainability and audit trails are valued as part of the prediction. Within this benchmark, LLM-SAR therefore plays its most natural role on toxicity related endpoints with classical structural alerts, such as AMES mutagenicity, Tox21 NR/SR assays and KCNH2/hERG cardiotoxicity, and on selected ADME related liabilities such as PXR/NR1I2 activation. It can also provide interpretable chemotype priors for phenotypic bioactivity endpoints, including nitroimidazole like motifs for anti-TB H37Rv activity or quinoline like and cationic features for antimalaria Pf. 3D7/Dd2 activity. By contrast, it is weaker for high throughput target based bioactivity prediction (e.g.\ EGFR kinase IC50) and for continuous ADME endpoints (Caco2, lipophilicity, solubility), where supervised regressors fit local SAR more efficiently than rule based scoring.

\paragraph{Two LLM origins, one evaluation protocol.}
The two LLM-SAR labels denote two rule libraries generated by different frontier large language models and evaluated under the same framework. GPT5.5-SAR is a GPT5.5 generated library of regex oriented SMILES rules with an associated scoring rubric. Opus4.7-SAR is an Opus 4.7 generated library of RDKit SMARTS patterns with explicit endpoint priors and a separate scoring rubric. Both libraries encode molecular structure, endpoint biology and medicinal chemistry reasoning, but both were frozen at generation time. Subsequent held out scoring is fully local: rule application is deterministic, knowledge variants are constructed exclusively from training fold molecules, and no test molecule is ever sent to a live LLM endpoint. Differences between the two labels therefore reflect (a) differences in the generated SAR rules and endpoint priors, and (b) differences in the rule grammar (regex on canonical SMILES vs.\ RDKit SMARTS).

\begin{figure}[!htbp]
\centering
\includegraphics[width=\linewidth]{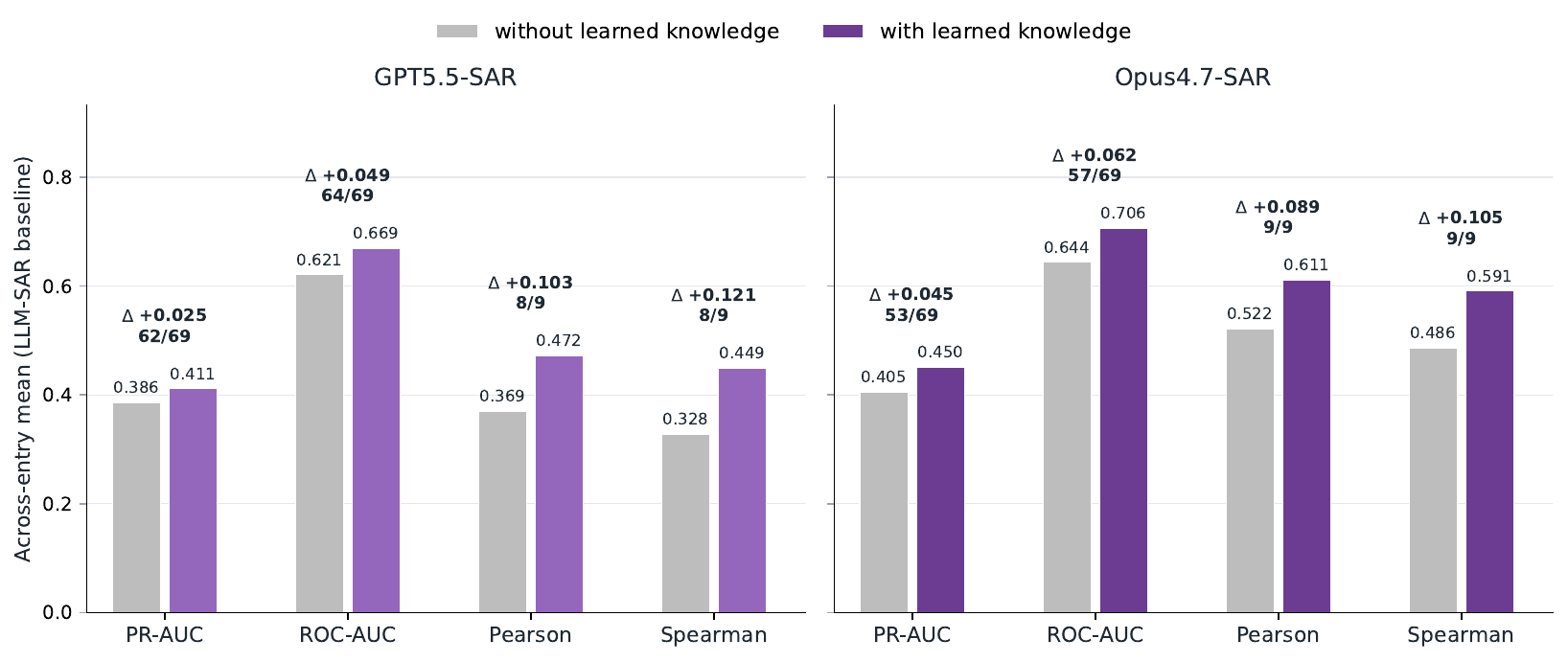}
\caption{Effect of SAR knowledge from training folds on LLM-SAR mean performance, shown as paired before / after bars per family across four higher is better metrics: PR-AUC, ROC-AUC, Pearson and Spearman. PR-AUC and ROC-AUC are computed over classification entries; Pearson and Spearman are computed over regression entries. The left panel reports GPT5.5-SAR; the right panel reports Opus4.7-SAR. Each pair of bars compares the across-entry mean without training fold knowledge (grey) and with training fold knowledge (coloured), computed over split level entries where both knowledge variants are available. The annotation above each pair gives the across-entry mean change $\Delta$ together with the count of entries improved by knowledge over the total number of comparable entries.}
\label{fig:llm-sar}
\end{figure}

\paragraph{Effect of SAR knowledge from training folds.}
Adding SAR knowledge from training folds improves many GPT5.5-SAR and Opus4.7-SAR metrics across classification discrimination and regression correlation views, although the direction and magnitude depend on endpoint class and metric (Fig.~\ref{fig:llm-sar}). Across the comparable split level classification entries, mean PR-AUC improves by approximately +0.025 for GPT5.5-SAR and +0.045 for Opus4.7-SAR; mean ROC-AUC improves by approximately +0.049 and +0.062, respectively. Across the regression entries only, Pearson correlation improves by +0.103 and +0.089, and Spearman correlation improves by +0.121 and +0.105. For anti-TB H37Rv and antimalaria Pf. 3D7/Dd2 cellular activity, mean PR-AUC increases by +0.027 for GPT5.5-SAR and +0.049 for Opus4.7-SAR, whereas mean ROC-AUC increases by +0.064 and +0.112. In classification, learned knowledge can improve ranking while shifting the operating point, which makes class specific precision and recall necessary for interpretation. The Opus4.7-SAR uplift is generally larger than the GPT5.5-SAR uplift on the same task, consistent with a richer SMARTS based rule grammar absorbing more endpoint specific structural information from the training fold.

\paragraph{Head to head behaviour.}
Under matched knowledge conditions in the split level Figure~\ref{fig:llm-sar} data, Opus4.7-SAR outperforms GPT5.5-SAR on PR-AUC for 58 of 69 comparable classification entries with knowledge and 49 of 69 without knowledge; for ROC-AUC the corresponding counts are 56 of 69 and 44 of 69. On the regression entries, Opus4.7-SAR is higher than GPT5.5-SAR on both Pearson and Spearman for all 9 comparable entries, with and without learned knowledge. The two LLM origins therefore behave as a controlled contrast: the same evaluation protocol, two rule grammars, two reasoning trails. The result favours the SMARTS and priors origin in the majority of entries, but neither LLM-SAR family becomes a substitute for the supervised classical ML/GNN/sequence predictors at the family winner level, where LLM-SAR wins three task and metric comparisons out of 156.

\paragraph{Practical value.}
LLM-SAR is most useful as an automated medicinal chemistry hypothesis generator and explanation layer, consistent with the broader motivation for explainable AI in drug discovery \citep{jimenezluna2020xai}. Without explicit assay mechanism prompts, training fold rule induction recovers recognisable medicinal chemistry motifs: nitro and polyaromatic alerts for mutagenicity, azole / lipophilic motifs for aromatase, nitroimidazole like patterns for anti-TB H37Rv activity, and quinoline like or cationic features for antimalaria Pf. 3D7/Dd2 activity. The complete rule listings induced from training folds for all tasks and folds are provided as CSV files rather than as typeset supplementary tables; representative examples are summarised in Table~\ref{tab:sar-rules}.

\paragraph{Bottom line.}
LLM-SAR is useful for rule generation, endpoint interpretation, mechanistic hypothesis formation and explainability when interacting with medicinal chemists. It is not a substitute for supervised molecular predictors when sufficient labelled data are available. The most defensible deployment pattern is to use LLM-SAR as a hypothesis and explanation layer on top of a supervised predictor, rather than as a standalone scorer.

\section{Discussion: Model Task Fit Over Model Scale}

The benchmark supports an evidence based interpretation: model scale is one variable among many, not a reliable ordering principle. The data instead point to recurring associations between endpoint structure, split difficulty, molecular representation and inductive bias. The strongest result is not that small models always win. It is that the simple rule that ``larger models should win'' fails across chemically separated folds, random folds, Murcko scaffold folds, liability endpoints and target activity endpoints.

\paragraph{Small models remain strong baselines.}
Tree ensembles, such as ExtraTrees or Random Forest variants using ECFP4, ECFP6 or RDKit descriptor representations, remain strong classical ML baselines. These models pair chemically meaningful fixed representations with stable supervised learning, and they exploit local SAR, substructure alerts and physicochemical patterns effectively. Their performance across ADME related, toxicity related and bioactivity related endpoint classes shows that compact fingerprint and descriptor based models remain difficult to displace, especially when endpoints are dominated by local analogue relationships, modest data size, label noise or class imbalance.

\paragraph{Graph and sequence models help selectively.}
They do not dominate the benchmark as a whole. Under the primary optimal held-out readout, they become more competitive in selected chemically harder splits. The fixed final-window sensitivity analysis shows that some strict first-place assignments for GNN and sequence models are readout-sensitive, whereas classical ML is unchanged by this choice and remains the most stable leading family. GIN and Ligandformer provide graph based inductive biases for molecular topology and substructure context, whereas ChemBERTa, ChemBERTa2 and MoLFormer provide pretrained SMILES representations. Their gains are therefore best interpreted as endpoint and split dependent complementarity, not as evidence that larger or pretrained models are universally superior.

\paragraph{LLM-SAR as reasoning rather than standalone prediction.}
LLM-SAR baselines are most informative when they summarize training set SAR, expose structural alerts and generate interpretable rules. Knowledge derived from training folds improves many SAR metrics, but the resulting predictors remain less competitive than supervised classical ML/GNN/sequence models on most primary metrics. This supports a complementary role for LLMs: reasoning support rather than default replacement.

\paragraph{Statistical support is stronger at the family level.}
The paired bootstrap analysis cautions against over reading first place ranks. The benchmark therefore supports family level statements more strongly than endpoint level claims that a single individual model is decisively superior.

\paragraph{Readout sensitivity limits strict winner claims.}
The fixed final-window sensitivity analysis reinforces this caution. In the idealized optimal held-out readout, GNN and sequence models can be competitive in selected settings, whereas the fixed final-window readout shows that classical ML is the most stable leading family. Strict winner shares should therefore be read as readout-sensitive summaries rather than as decisive rankings.

\paragraph{Evaluation protocol shapes apparent scaling.}
Random like splits, scaffold splits and chemical cluster splits can produce different model rankings. Without precise protocol reporting, claims about larger models winning or losing are difficult to interpret.

The four questions posed in the Introduction therefore receive answers that are not monotonic:
\begin{itemize}
\item \textbf{Does parameter count predict endpoint level winning?} No. If parameter count were sufficient, pretrained sequence models and SAR rule baselines would dominate more entries. Instead, classical ML remains the largest winner family overall, while GNN and sequence models are competitive in selected harder split settings under the primary optimal held-out readout.
\item \textbf{Does pretraining corpus size predict winning?} Selectively. SMILES pretraining helps when token level chemical regularities transfer to the endpoint, but it does not replace assay specific labels.
\item \textbf{Which representation provides the decisive inductive bias?} The answer is task dependent. ECFP tree ensembles are strong in local SAR and anti-TB H37Rv regimes, whereas graph models win when topology and learned message passing align with endpoint biology.
\item \textbf{Where does explicit SAR reasoning help or mislead?} LLM-SAR helps most as an interpretation and hypothesis layer. It can mislead as a decision rule when fold specific rules become too conservative, too broad or poorly calibrated.
\end{itemize}

\section{Limitations}

Several limitations affect the scope of the evidence. First, the benchmark remains an empirical comparison over selected endpoints rather than a universal theory of molecular scaling. Second, paired bootstrap quantifies uncertainty over the available held out prediction pools, but it does not replace prospective validation. Third, top K enrichment and calibration are included as supporting analyses, yet the main claims are still driven by rank based and error based metrics. Fourth, the primary infectious disease conclusions use anti-TB H37Rv and antimalaria Pf. 3D7/Dd2 public datasets; restricted institutional infectious disease datasets are reported separately as supplementary restricted data analyses (Supplementary Note~5) because their raw structural records require source provenance checks and publication permissions before public redistribution.

\section{Conclusion}

This benchmark does not support a simple scale centred narrative for molecular property and activity prediction. Across ADME related, toxicity related and bioactivity related endpoint classes, the differences among trained model families often depend on both endpoint and split. Classical ML remains the most stable leading family, whereas GNN and pretrained molecular sequence models provide endpoint and split specific gains rather than a universal scaling advantage. Larger or more general models therefore do not provide a universal predictive advantage. Their value is better framed as complementary: they can support zero shot reasoning, SAR interpretation and hypothesis generation, but the evidence does not support treating them as substitutes for supervised predictors when labelled molecular data are available. The central implication is that model scale is a weak explanatory variable unless it is interpreted through molecular representation, inductive bias, data regime, endpoint biology and validation design.

\section*{Code and Data Availability}

The code and structured data required to reproduce the data splitting, model training, prediction post processing, statistical analyses, tables and figures are provided in the accompanying reproducibility package. Fold assignments, table data, figure data, checksum manifests and derived summaries are provided with the package. Public ADME, toxicity and bioactivity datasets are available from the sources cited in the manuscript, including KCNH2/hERG inhibition, DRD2 receptor binding, EGFR kinase inhibition, PXR/NR1I2 activation, anti-TB H37Rv activity and antimalaria Pf. 3D7/Dd2 activity records. Restricted institutional infectious disease datasets are not required to reproduce the primary conclusions and are not publicly redistributed in raw structural form; derived aggregate summaries may be made available subject to institutional approval.

\bibliographystyle{unsrtnat}
\bibliography{references}

\clearpage
\section*{Tables}

\input{tables/table_endpoint_classification.tex}

\input{tables/table_dataset_sizes.tex}

\input{tables/table_family_winner_summary.tex}

\input{tables/table_structure_adme_prauc_rocauc_model_matrix.tex}
\input{tables/table_structure_adme_mae_pearson_model_matrix.tex}
\input{tables/table_structure_toxicity_general_prauc_rocauc_model_matrix.tex}
\input{tables/table_structure_toxicity_nr_prauc_rocauc_model_matrix.tex}
\input{tables/table_structure_toxicity_sr_prauc_rocauc_model_matrix.tex}
\input{tables/table_structure_bioactivity_prauc_rocauc_model_matrix.tex}
\clearpage

\input{tables/table_split_realism_drops.tex}

\input{tables/table_topk_best_ef1.tex}

\input{tables/table_data_regime_best.tex}

\input{tables/table_data_regime_first_crossover.tex}

\input{tables/table_prevalence_leaders.tex}
\clearpage

\input{tables/table_paired_bootstrap_headlines.tex}

\input{tables/table_calibration_best_by_class.tex}

\input{tables/table_sar_rule_examples.tex}

\end{document}

%% file: tables/table_endpoint_classification.tex
\begin{table}[H]
\centering
\small
\caption{Endpoint class definitions used in the manuscript. The benchmark is grouped into three main pharmacological classes while retaining endpoint subtypes for biological specificity. Dataset endpoints (n) counts endpoint, label and source units used in the benchmark denominator.}
\label{tab:endpoint-classes}
\resizebox{\linewidth}{!}{%
\begin{tabular}{p{0.16\linewidth}p{0.18\linewidth}p{0.07\linewidth}p{0.28\linewidth}p{0.26\linewidth}}
\toprule
Endpoint class & Endpoint subtype & Dataset endpoints (n) & Endpoints & Interpretation \\
\midrule
ADME related & Permeability / distribution & 2 & Caco2 permeability; blood brain barrier penetration & Absorption and distribution endpoints that depend on permeability, polarity, ionization, lipophilicity and transporter related effects. \\
ADME related & Metabolism / DDI liability & 2 & CYP3A4 inhibition; PXR/NR1I2 activation & Metabolic liability and drug interaction risk through CYP inhibition or xenobiotic receptor activation. \\
ADME related & Physicochemical ADME property & 2 & Lipophilicity; aqueous solubility & Physicochemical properties that shape solubility, permeability, distribution and developability. \\
Toxicity related & Mutagenicity / hepatotoxicity & 2 & AMES mutagenicity; DILI & Classical toxicity liabilities covering genotoxicity and drug induced liver injury. \\
Toxicity related & Cardiotoxicity / safety pharmacology & 2 & hERG liability, TDC binary label; KCNH2/hERG inhibition, ChEMBL IC$_{50}$ label & Cardiac IKr channel inhibition and related cardiotoxicity liability. \\
Toxicity related & Tox21 pathway toxicity & 12 & 12 Tox21 nuclear receptor and stress response assays & Pathway level receptor activation and cellular stress response toxicity assays. \\
Toxicity related & Off target CNS safety liability & 1 & DRD2 receptor binding Ki & CNS off target pharmacology with safety liability relevance. \\
Bioactivity related & Target based bioactivity & 1 & EGFR kinase inhibition IC50 & Single target kinase activity benchmark. \\
Bioactivity related & Phenotypic pathogen activity & 2 & anti-TB H37Rv cellular activity at 1 uM; antimalaria Pf. 3D7/Dd2 cellular activity at 100 nM & Pathogen growth inhibition in bacterial or parasite phenotypic assays. \\
\bottomrule
\end{tabular}
}
\end{table}

%% file: tables/table_dataset_sizes.tex
\begin{table}[H]
\centering
\small
\caption{Dataset sizes for the 26 endpoints in the benchmark. ``Molecules'' is the deduplicated unique compound count after canonical SMILES standardisation and conflict resolution within each endpoint. ``Positive rate'' is the active fraction for classification endpoints; the three regression endpoints (Caco2 permeability, lipophilicity, aqueous solubility) report mean and range of the continuous target instead. The 12 Tox21 assays carry per assay missing labels, so the same compound can appear in different assays with different label-presence patterns; per-assay molecule counts are reported. Endpoint specific molecule counts sum to 165{,}541 endpoint level compound label records across 23 classification and 3 regression endpoints.}
\label{tab:dataset-sizes}
\resizebox{\linewidth}{!}{%
\begin{tabular}{lllrrll}
\toprule
Endpoint class & Endpoint subtype & Endpoint & Molecules & Positive rate or mean & Source & Reference \\
\midrule
ADME related & Permeability / distribution & Caco2 permeability & 906 & regression (log Papp) & TDC ADMET & \cite{tdc2021} \\
ADME related & Permeability / distribution & Blood brain barrier penetration & 1{,}965 & 76.3\% & TDC ADMET & \cite{tdc2021} \\
ADME related & Metabolism / DDI liability & CYP3A4 inhibition & 12{,}328 & 41.5\% & TDC ADMET & \cite{tdc2021} \\
ADME related & Metabolism / DDI liability & PXR/NR1I2 activation & 1{,}175 & 77.7\% & ChEMBL EC50/AC50 & \cite{mendez2019bioassay,kliewer1998pxr} \\
ADME related & Physicochemical ADME property & Lipophilicity & 4{,}200 & regression (logD$_{7.4}$) & TDC ADMET & \cite{tdc2021} \\
ADME related & Physicochemical ADME property & Aqueous solubility & 9{,}982 & regression (log S) & TDC ADMET & \cite{tdc2021} \\
\midrule
Toxicity related & Mutagenicity / hepatotoxicity & AMES & 7{,}255 & 54.5\% & TDC ADMET & \cite{tdc2021} \\
Toxicity related & Mutagenicity / hepatotoxicity & DILI & 475 & 49.7\% & TDC ADMET & \cite{tdc2021} \\
Toxicity related & Cardiotoxicity / safety pharmacology & hERG liability, TDC binary label & 645 & 68.8\% & TDC ADMET & \cite{tdc2021,wang2016herg} \\
Toxicity related & Cardiotoxicity / safety pharmacology & KCNH2/hERG inhibition, ChEMBL IC$_{50}$ label & 9{,}516 & 21.8\% & ChEMBL IC$_{50}$ & \cite{mendez2019bioassay,sanguinetti1995herg} \\
Toxicity related & Tox21 pathway toxicity & tox21\_NR-AR & 7{,}265 & 4.3\% & Tox21 & \cite{tox21_challenge2015,tox21_qsar2016} \\
Toxicity related & Tox21 pathway toxicity & tox21\_NR-AR-LBD & 6{,}758 & 3.5\% & Tox21 & \cite{tox21_challenge2015,tox21_qsar2016} \\
Toxicity related & Tox21 pathway toxicity & tox21\_NR-AhR & 6{,}549 & 11.7\% & Tox21 & \cite{tox21_challenge2015,tox21_qsar2016} \\
Toxicity related & Tox21 pathway toxicity & tox21\_NR-Aromatase & 5{,}821 & 5.2\% & Tox21 & \cite{tox21_challenge2015,tox21_qsar2016} \\
Toxicity related & Tox21 pathway toxicity & tox21\_NR-ER & 6{,}193 & 12.8\% & Tox21 & \cite{tox21_challenge2015,tox21_qsar2016} \\
Toxicity related & Tox21 pathway toxicity & tox21\_NR-ER-LBD & 6{,}955 & 5.0\% & Tox21 & \cite{tox21_challenge2015,tox21_qsar2016} \\
Toxicity related & Tox21 pathway toxicity & tox21\_NR-PPAR-$\gamma$ & 6{,}450 & 2.9\% & Tox21 & \cite{tox21_challenge2015,tox21_qsar2016} \\
Toxicity related & Tox21 pathway toxicity & tox21\_SR-ARE & 5{,}832 & 16.2\% & Tox21 & \cite{tox21_challenge2015,tox21_qsar2016} \\
Toxicity related & Tox21 pathway toxicity & tox21\_SR-ATAD5 & 7{,}072 & 3.7\% & Tox21 & \cite{tox21_challenge2015,tox21_qsar2016} \\
Toxicity related & Tox21 pathway toxicity & tox21\_SR-HSE & 6{,}467 & 5.8\% & Tox21 & \cite{tox21_challenge2015,tox21_qsar2016} \\
Toxicity related & Tox21 pathway toxicity & tox21\_SR-MMP & 5{,}810 & 15.8\% & Tox21 & \cite{tox21_challenge2015,tox21_qsar2016} \\
Toxicity related & Tox21 pathway toxicity & tox21\_SR-p53 & 6{,}774 & 6.2\% & Tox21 & \cite{tox21_challenge2015,tox21_qsar2016} \\
Toxicity related & Off target CNS safety liability & DRD2 receptor binding (Ki) & 7{,}956 & 88.9\% & ChEMBL Ki (CHEMBL217) & \cite{mendez2019bioassay,deGreef2011d2occupancy} \\
\midrule
Bioactivity related & Target based bioactivity & EGFR kinase inhibition (IC50) & 11{,}301 & 79.0\% & ChEMBL IC50 (CHEMBL203) & \cite{mendez2019bioassay} \\
Bioactivity related & Phenotypic pathogen activity & anti-TB H37Rv at 1~$\mu$M & 15{,}467 & 12.8\% & public HTS and literature & \cite{ballell2013opensourceTB,rebollo2015opensourceTB,lane2018mtbml,eke2025mlsmrTB,chitale2022h37rv} \\
Bioactivity related & Phenotypic pathogen activity & antimalaria Pf. 3D7/Dd2 at 100~nM & 4{,}424 & 45.2\% & public cellular activity records & \cite{iwanaga2022pfresistance} \\
\bottomrule
\end{tabular}
}
\end{table}

%% file: tables/table_family_winner_summary.tex
\begin{table}[H]
\centering
\small
\caption{Fold mean task and metric family winners by split and endpoint class under the primary optimal held-out readout. Classification endpoints contribute PR-AUC and ROC-AUC winners; regression endpoints contribute MAE and Pearson winners. MAE uses the lowest value as the winner; all other metrics use the highest value.}
\label{tab:family-winner-summary}
\resizebox{\linewidth}{!}{%
\begin{tabular}{llrrrrrr}
\toprule
Split & Endpoint class & Task and metric entries & ML & GNN & Sequence & LLM-SAR & Total \\
\midrule
Random & ADME related & 12 & 5 & 0 & 7 & 0 & 12 \\
Random & Toxicity related & 34 & 28 & 2 & 3 & 1 & 34 \\
Random & Bioactivity related & 6 & 5 & 0 & 1 & 0 & 6 \\
\midrule
Murcko scaffold & ADME related & 12 & 5 & 2 & 5 & 0 & 12 \\
Murcko scaffold & Toxicity related & 34 & 15 & 10 & 8 & 1 & 34 \\
Murcko scaffold & Bioactivity related & 6 & 5 & 1 & 0 & 0 & 6 \\
\midrule
Structure separated & ADME related & 12 & 1 & 4 & 7 & 0 & 12 \\
Structure separated & Toxicity related & 34 & 8 & 12 & 13 & 1 & 34 \\
Structure separated & Bioactivity related & 6 & 2 & 3 & 1 & 0 & 6 \\
\midrule
Total & All endpoint classes & 156 & 74 & 34 & 45 & 3 & 156 \\
\bottomrule
\end{tabular}
}
\end{table}

%% file: tables/table_structure_adme_prauc_rocauc_model_matrix.tex
\begin{table*}[t]
\centering
\caption{Structure separated CV model-by-endpoint matrix for ADME-related classification endpoints. Each endpoint entry reports PR-AUC / ROC-AUC. Rankings are computed independently for each metric within each endpoint: the first ranked value is bold and underlined, and the second and third ranked values are underlined.}
\label{tab:table-structure-adme-prauc-rocauc-model-matrix}
\tiny
\setlength{\tabcolsep}{2pt}
\resizebox{\textwidth}{!}{%
\begin{tabular}{llccc}
\toprule
Family & Model & BBB & CYP3A4 & PXR/NR1I2 \\
\midrule
ML & ExtraTrees(ECFP4) & 0.753 / 0.676 & 0.724 / 0.793 & 0.803 / 0.733 \\
ML & ExtraTrees(ECFP6) & 0.748 / 0.669 & 0.711 / 0.785 & 0.797 / 0.722 \\
ML & ExtraTrees(RDKit desc.) & 0.765 / 0.700 & 0.711 / 0.787 & \underline{0.839} / \underline{0.784} \\
ML & ExtraTrees(MACCS) & 0.753 / 0.649 & 0.683 / 0.760 & 0.774 / 0.722 \\
ML & RF(ECFP4) & 0.756 / 0.697 & 0.728 / 0.795 & 0.836 / 0.754 \\
ML & RF(ECFP6) & 0.743 / 0.645 & 0.711 / 0.785 & 0.802 / 0.716 \\
ML & RF(RDKit desc.) & 0.752 / 0.650 & 0.714 / 0.789 & 0.814 / 0.754 \\
ML & RF(MACCS) & 0.761 / 0.673 & 0.679 / 0.757 & 0.777 / 0.729 \\
ML & GBDT(ECFP4) & 0.739 / 0.628 & 0.724 / 0.783 & 0.784 / 0.721 \\
ML & GBDT(ECFP6) & 0.728 / 0.626 & 0.724 / 0.782 & 0.825 / 0.730 \\
ML & GBDT(RDKit desc.) & 0.746 / 0.631 & 0.731 / 0.804 & 0.793 / 0.758 \\
ML & GBDT(MACCS) & 0.751 / 0.685 & 0.705 / 0.776 & 0.765 / 0.695 \\
ML & LR(ECFP4) & 0.757 / 0.624 & 0.650 / 0.729 & 0.759 / 0.686 \\
ML & LR(ECFP6) & 0.730 / 0.596 & 0.622 / 0.703 & 0.682 / 0.560 \\
ML & LR(RDKit desc.) & 0.753 / 0.685 & 0.702 / 0.783 & 0.760 / 0.712 \\
ML & LR(MACCS) & 0.742 / 0.635 & 0.675 / 0.753 & 0.804 / 0.721 \\
\midrule
GNN & Ligandformer & \underline{0.784} / \underline{0.791} & 0.760 / 0.815 & \textbf{\underline{0.883}} / \textbf{\underline{0.821}} \\
GNN & GAT & 0.778 / 0.747 & \underline{0.770} / \underline{0.827} & 0.838 / 0.750 \\
GNN & GCN & 0.775 / 0.748 & 0.766 / 0.819 & 0.805 / 0.680 \\
GNN & GIN & \textbf{\underline{0.794}} / \underline{0.770} & \underline{0.771} / \textbf{\underline{0.834}} & 0.824 / 0.717 \\
\midrule
Sequence & ChemBERTa & 0.762 / 0.681 & 0.713 / 0.787 & 0.710 / 0.637 \\
Sequence & ChemBERTa2 & 0.775 / 0.735 & 0.746 / 0.804 & 0.827 / 0.738 \\
Sequence & MoLFormer & \underline{0.782} / \textbf{\underline{0.799}} & \textbf{\underline{0.773}} / \underline{0.828} & \underline{0.856} / \underline{0.809} \\
\midrule
LLM-SAR & GPT5.5-SAR & 0.732 / 0.585 & 0.578 / 0.684 & 0.693 / 0.567 \\
LLM-SAR & GPT5.5-SAR + knowledge & 0.731 / 0.587 & 0.586 / 0.698 & 0.692 / 0.571 \\
LLM-SAR & Opus4.7-SAR & 0.723 / 0.592 & 0.591 / 0.690 & 0.686 / 0.582 \\
LLM-SAR & Opus4.7-SAR + knowledge & 0.742 / 0.619 & 0.636 / 0.727 & 0.752 / 0.537 \\
\bottomrule
\end{tabular}
}
\end{table*}

%% file: tables/table_structure_adme_mae_pearson_model_matrix.tex
\begin{table*}[t]
\centering
\caption{Structure separated CV model-by-endpoint matrix for ADME-related regression endpoints. Each endpoint entry reports MAE / Pearson. Lower MAE is better; higher Pearson is better. Rankings are computed independently for each metric within each endpoint: the first ranked value is bold and underlined, and the second and third ranked values are underlined.}
\label{tab:table-structure-adme-mae-pearson-model-matrix}
\tiny
\setlength{\tabcolsep}{2pt}
\resizebox{\textwidth}{!}{%
\begin{tabular}{llccc}
\toprule
Family & Model & Caco2 & Lipophilicity & Solubility \\
\midrule
ML & ExtraTrees(ECFP4) & 0.823 / 0.222 & 1.059 / 0.335 & 1.692 / 0.444 \\
ML & ExtraTrees(ECFP6) & 0.743 / 0.286 & 1.128 / 0.219 & 1.798 / 0.398 \\
ML & ExtraTrees(RDKit desc.) & \underline{0.401} / \underline{0.661} & 0.626 / 0.711 & \underline{0.837} / \textbf{\underline{0.840}} \\
ML & ExtraTrees(MACCS) & 0.545 / 0.465 & 1.090 / 0.359 & 1.586 / 0.541 \\
ML & RF(ECFP4) & 0.589 / 0.450 & 0.815 / 0.496 & 1.416 / 0.583 \\
ML & RF(ECFP6) & 0.545 / 0.369 & 0.808 / 0.492 & 1.434 / 0.575 \\
ML & RF(RDKit desc.) & \underline{0.407} / 0.629 & 0.658 / 0.683 & 0.862 / \underline{0.829} \\
ML & RF(MACCS) & 0.456 / 0.611 & 0.775 / 0.574 & 1.295 / 0.674 \\
ML & GBDT(ECFP4) & 0.580 / 0.474 & 0.794 / 0.486 & 1.393 / 0.619 \\
ML & GBDT(ECFP6) & 0.587 / 0.314 & 0.813 / 0.464 & 1.405 / 0.613 \\
ML & GBDT(RDKit desc.) & 0.413 / \underline{0.640} & 0.637 / 0.695 & \underline{0.854} / 0.828 \\
ML & GBDT(MACCS) & 0.423 / 0.628 & 0.794 / 0.541 & 1.281 / 0.681 \\
ML & Ridge(ECFP4) & 0.587 / 0.395 & 1.457 / 0.343 & 1.509 / 0.572 \\
ML & Ridge(ECFP6) & 0.555 / 0.294 & 1.439 / 0.355 & 1.491 / 0.577 \\
ML & Ridge(RDKit desc.) & 0.487 / 0.635 & 0.756 / 0.615 & 1.242 / 0.628 \\
ML & Ridge(MACCS) & 0.531 / 0.585 & 0.869 / 0.445 & 1.430 / 0.629 \\
\midrule
GNN & Ligandformer & 0.941 / 0.334 & \underline{0.541} / \underline{0.794} & 0.957 / 0.797 \\
GNN & GAT & 1.253 / 0.157 & 0.606 / 0.734 & 1.325 / 0.660 \\
GNN & GCN & 0.927 / 0.275 & 0.628 / 0.694 & 0.958 / 0.795 \\
GNN & GIN & 0.794 / 0.302 & \underline{0.559} / \underline{0.777} & 0.920 / 0.810 \\
\midrule
Sequence & ChemBERTa & 0.694 / 0.257 & 0.700 / 0.586 & 0.917 / 0.811 \\
Sequence & ChemBERTa2 & 0.514 / 0.484 & 0.608 / 0.672 & 0.879 / 0.808 \\
Sequence & MoLFormer & \textbf{\underline{0.399}} / \textbf{\underline{0.663}} & \textbf{\underline{0.509}} / \textbf{\underline{0.800}} & \textbf{\underline{0.817}} / \underline{0.837} \\
\midrule
LLM-SAR & GPT5.5-SAR & 0.640 / 0.231 & 1.492 / 0.380 & 2.344 / 0.502 \\
LLM-SAR & GPT5.5-SAR + knowledge & 0.625 / 0.430 & 2.074 / 0.342 & 1.885 / 0.524 \\
LLM-SAR & Opus4.7-SAR & 0.926 / 0.459 & 1.526 / 0.386 & 1.878 / 0.610 \\
LLM-SAR & Opus4.7-SAR + knowledge & 1.209 / 0.522 & 1.457 / 0.505 & 2.015 / 0.656 \\
\bottomrule
\end{tabular}
}
\end{table*}

%% file: tables/table_structure_toxicity_general_prauc_rocauc_model_matrix.tex
\begin{table*}[t]
\centering
\caption{Structure separated CV model-by-endpoint matrix for general toxicity and safety liability endpoints. Each endpoint entry reports PR-AUC / ROC-AUC. Rankings are computed independently for each metric within each endpoint: the first ranked value is bold and underlined, and the second and third ranked values are underlined.}
\label{tab:table-structure-toxicity-general-prauc-rocauc-model-matrix}
\tiny
\setlength{\tabcolsep}{2pt}
\resizebox{\textwidth}{!}{%
\begin{tabular}{llccccc}
\toprule
Family & Model & AMES & DILI & TDC hERG & hERG/KCNH2 & DRD2 \\
\midrule
ML & ExtraTrees(ECFP4) & 0.813 / 0.769 & 0.793 / 0.792 & 0.736 / 0.612 & 0.693 / 0.841 & 0.972 / 0.773 \\
ML & ExtraTrees(ECFP6) & 0.802 / 0.758 & 0.765 / 0.774 & 0.828 / 0.754 & 0.683 / 0.836 & \underline{0.972} / \textbf{\underline{0.838}} \\
ML & ExtraTrees(RDKit desc.) & \underline{0.816} / 0.782 & 0.821 / 0.837 & 0.878 / \underline{0.846} & \underline{0.714} / \underline{0.851} & 0.964 / 0.738 \\
ML & ExtraTrees(MACCS) & 0.796 / 0.754 & \textbf{\underline{0.835}} / \underline{0.850} & \underline{0.901} / 0.794 & 0.664 / 0.832 & 0.960 / 0.702 \\
ML & RF(ECFP4) & 0.808 / 0.766 & 0.803 / 0.798 & 0.750 / 0.644 & 0.683 / 0.835 & 0.971 / 0.793 \\
ML & RF(ECFP6) & 0.800 / 0.754 & 0.789 / 0.792 & 0.826 / 0.754 & 0.677 / 0.833 & 0.972 / 0.805 \\
ML & RF(RDKit desc.) & 0.805 / 0.777 & \underline{0.823} / 0.830 & 0.791 / 0.723 & 0.679 / 0.838 & 0.961 / 0.714 \\
ML & RF(MACCS) & 0.802 / 0.750 & \underline{0.833} / \underline{0.850} & 0.897 / 0.779 & 0.674 / 0.835 & 0.962 / 0.698 \\
ML & GBDT(ECFP4) & 0.781 / 0.744 & 0.736 / 0.746 & 0.749 / 0.692 & 0.599 / 0.767 & 0.966 / 0.737 \\
ML & GBDT(ECFP6) & 0.766 / 0.732 & 0.693 / 0.715 & 0.814 / 0.733 & 0.597 / 0.776 & 0.967 / 0.782 \\
ML & GBDT(RDKit desc.) & 0.779 / 0.748 & 0.775 / 0.793 & 0.758 / 0.619 & 0.625 / 0.797 & 0.957 / 0.701 \\
ML & GBDT(MACCS) & 0.784 / 0.752 & 0.778 / 0.808 & 0.882 / 0.759 & 0.591 / 0.793 & 0.958 / 0.725 \\
ML & LR(ECFP4) & 0.667 / 0.655 & 0.722 / 0.747 & 0.723 / 0.653 & 0.537 / 0.765 & 0.960 / 0.724 \\
ML & LR(ECFP6) & 0.688 / 0.652 & 0.693 / 0.721 & 0.876 / 0.719 & 0.525 / 0.763 & 0.958 / 0.715 \\
ML & LR(RDKit desc.) & 0.745 / 0.734 & 0.752 / 0.787 & 0.768 / 0.730 & 0.504 / 0.756 & 0.959 / 0.681 \\
ML & LR(MACCS) & 0.745 / 0.721 & 0.740 / 0.744 & 0.713 / 0.590 & 0.476 / 0.727 & 0.955 / 0.679 \\
\midrule
GNN & Ligandformer & 0.807 / \underline{0.790} & 0.810 / 0.832 & \underline{0.916} / 0.815 & 0.702 / 0.846 & \textbf{\underline{0.973}} / \underline{0.835} \\
GNN & GAT & 0.804 / \textbf{\underline{0.791}} & 0.801 / 0.834 & 0.890 / 0.784 & 0.702 / 0.837 & 0.968 / \underline{0.817} \\
GNN & GCN & 0.800 / 0.784 & 0.767 / 0.802 & 0.876 / 0.779 & 0.697 / \underline{0.851} & 0.968 / 0.802 \\
GNN & GIN & 0.801 / 0.782 & 0.775 / 0.829 & 0.847 / 0.725 & \underline{0.710} / 0.849 & 0.969 / 0.805 \\
\midrule
Sequence & ChemBERTa & 0.781 / 0.738 & 0.729 / 0.757 & 0.880 / 0.772 & 0.627 / 0.795 & 0.962 / 0.708 \\
Sequence & ChemBERTa2 & \underline{0.814} / 0.785 & 0.793 / 0.818 & 0.868 / \underline{0.824} & 0.690 / 0.831 & 0.970 / 0.815 \\
Sequence & MoLFormer & \textbf{\underline{0.820}} / \underline{0.787} & 0.813 / \textbf{\underline{0.857}} & \textbf{\underline{0.940}} / \textbf{\underline{0.872}} & \textbf{\underline{0.719}} / \textbf{\underline{0.854}} & \underline{0.972} / 0.807 \\
\midrule
LLM-SAR & GPT5.5-SAR & 0.570 / 0.543 & 0.652 / 0.654 & 0.799 / 0.701 & 0.295 / 0.611 & 0.941 / 0.564 \\
LLM-SAR & GPT5.5-SAR + knowledge & 0.575 / 0.546 & 0.692 / 0.712 & 0.783 / 0.606 & 0.361 / 0.688 & 0.943 / 0.562 \\
LLM-SAR & Opus4.7-SAR & 0.702 / 0.669 & 0.534 / 0.518 & 0.768 / 0.648 & 0.379 / 0.702 & 0.939 / 0.531 \\
LLM-SAR & Opus4.7-SAR + knowledge & 0.671 / 0.663 & 0.656 / 0.702 & 0.762 / 0.624 & 0.413 / 0.712 & 0.951 / 0.649 \\
\bottomrule
\end{tabular}
}
\end{table*}

%% file: tables/table_structure_toxicity_nr_prauc_rocauc_model_matrix.tex
\begin{table*}[t]
\centering
\caption{Structure separated CV model-by-endpoint matrix for Tox21 nuclear receptor assays. Each endpoint entry reports PR-AUC / ROC-AUC. Rankings are computed independently for each metric within each endpoint: the first ranked value is bold and underlined, and the second and third ranked values are underlined.}
\label{tab:table-structure-toxicity-nr-prauc-rocauc-model-matrix}
\tiny
\setlength{\tabcolsep}{2pt}
\resizebox{\textwidth}{!}{%
\begin{tabular}{llccccccc}
\toprule
Family & Model & NR-AR & NR-AR-LBD & NR-AhR & NR-Aromatase & NR-ER & NR-ER-LBD & NR-PPAR-$\gamma$ \\
\midrule
ML & ExtraTrees(ECFP4) & 0.082 / 0.572 & 0.104 / 0.605 & \underline{0.338} / 0.802 & 0.139 / 0.659 & 0.242 / 0.594 & 0.316 / 0.755 & \underline{0.166} / 0.736 \\
ML & ExtraTrees(ECFP6) & 0.089 / 0.558 & 0.098 / 0.613 & 0.314 / 0.791 & 0.163 / 0.676 & 0.252 / 0.608 & 0.308 / 0.762 & 0.137 / 0.723 \\
ML & ExtraTrees(RDKit desc.) & 0.073 / 0.537 & \underline{0.147} / 0.630 & \underline{0.361} / 0.807 & 0.169 / 0.754 & 0.289 / 0.656 & \textbf{\underline{0.332}} / 0.753 & \textbf{\underline{0.172}} / 0.720 \\
ML & ExtraTrees(MACCS) & 0.071 / 0.496 & 0.090 / 0.578 & 0.274 / 0.765 & 0.114 / 0.678 & 0.252 / 0.627 & 0.261 / 0.704 & 0.112 / 0.708 \\
ML & RF(ECFP4) & 0.091 / 0.571 & 0.102 / 0.610 & 0.328 / 0.788 & 0.139 / 0.658 & 0.254 / 0.602 & \underline{0.326} / 0.748 & 0.164 / 0.707 \\
ML & RF(ECFP6) & 0.086 / 0.532 & 0.106 / 0.621 & 0.313 / 0.780 & 0.160 / 0.670 & 0.256 / 0.609 & 0.321 / \underline{0.768} & 0.141 / 0.686 \\
ML & RF(RDKit desc.) & 0.071 / 0.532 & \underline{0.150} / 0.640 & 0.326 / 0.784 & 0.151 / 0.733 & \underline{0.306} / \underline{0.665} & 0.323 / \underline{0.771} & 0.161 / 0.705 \\
ML & RF(MACCS) & 0.073 / 0.540 & 0.074 / 0.530 & 0.296 / 0.777 & 0.121 / 0.680 & 0.260 / 0.629 & 0.295 / 0.718 & 0.110 / 0.715 \\
ML & GBDT(ECFP4) & 0.086 / 0.539 & 0.093 / 0.558 & 0.322 / 0.783 & 0.102 / 0.556 & 0.227 / 0.552 & 0.293 / 0.703 & 0.099 / 0.606 \\
ML & GBDT(ECFP6) & 0.073 / 0.520 & 0.078 / 0.531 & 0.264 / 0.742 & 0.132 / 0.611 & 0.243 / 0.591 & 0.278 / 0.691 & 0.075 / 0.599 \\
ML & GBDT(RDKit desc.) & 0.073 / 0.516 & 0.096 / 0.545 & 0.334 / 0.772 & 0.130 / 0.695 & 0.270 / 0.619 & 0.287 / 0.710 & 0.157 / 0.713 \\
ML & GBDT(MACCS) & 0.068 / 0.497 & 0.066 / 0.391 & 0.286 / 0.779 & 0.119 / 0.652 & 0.251 / 0.618 & 0.239 / 0.660 & 0.099 / 0.700 \\
ML & LR(ECFP4) & 0.064 / 0.517 & 0.082 / 0.542 & 0.190 / 0.676 & 0.101 / 0.602 & 0.203 / 0.566 & 0.181 / 0.677 & 0.135 / 0.639 \\
ML & LR(ECFP6) & 0.070 / 0.490 & 0.077 / 0.537 & 0.171 / 0.622 & 0.074 / 0.541 & 0.165 / 0.521 & 0.191 / 0.644 & 0.106 / 0.666 \\
ML & LR(RDKit desc.) & 0.074 / 0.509 & 0.124 / 0.548 & 0.293 / 0.793 & 0.135 / 0.685 & 0.230 / 0.611 & 0.171 / 0.672 & 0.070 / 0.600 \\
ML & LR(MACCS) & 0.071 / 0.511 & 0.103 / 0.546 & 0.275 / 0.777 & 0.127 / 0.705 & 0.237 / 0.614 & 0.156 / 0.659 & 0.078 / 0.691 \\
\midrule
GNN & Ligandformer & \underline{0.120} / \underline{0.640} & \textbf{\underline{0.155}} / \textbf{\underline{0.688}} & 0.327 / 0.816 & \underline{0.219} / \underline{0.779} & \textbf{\underline{0.316}} / \textbf{\underline{0.669}} & 0.301 / \textbf{\underline{0.779}} & 0.123 / 0.771 \\
GNN & GAT & \underline{0.124} / \underline{0.628} & 0.108 / 0.643 & 0.314 / 0.815 & 0.167 / 0.741 & 0.289 / 0.648 & 0.269 / 0.748 & \underline{0.169} / \underline{0.773} \\
GNN & GCN & 0.094 / 0.619 & 0.094 / 0.663 & 0.317 / 0.815 & 0.178 / 0.751 & 0.284 / 0.650 & 0.241 / 0.748 & 0.127 / \underline{0.774} \\
GNN & GIN & 0.099 / 0.614 & 0.113 / \underline{0.675} & 0.301 / \underline{0.816} & 0.201 / 0.748 & 0.297 / 0.652 & 0.243 / 0.730 & 0.164 / \textbf{\underline{0.797}} \\
\midrule
Sequence & ChemBERTa & 0.118 / 0.600 & 0.138 / 0.671 & 0.311 / 0.808 & 0.186 / 0.763 & 0.256 / 0.629 & 0.260 / 0.731 & 0.088 / 0.695 \\
Sequence & ChemBERTa2 & 0.090 / 0.590 & 0.080 / 0.637 & \textbf{\underline{0.374}} / \textbf{\underline{0.835}} & 0.189 / \underline{0.767} & \underline{0.303} / \underline{0.668} & \underline{0.327} / 0.754 & 0.110 / 0.715 \\
Sequence & MoLFormer & \textbf{\underline{0.128}} / \textbf{\underline{0.674}} & 0.113 / \underline{0.681} & 0.333 / \underline{0.821} & 0.191 / \textbf{\underline{0.781}} & 0.294 / 0.661 & 0.279 / 0.748 & 0.104 / 0.741 \\
\midrule
LLM-SAR & GPT5.5-SAR & 0.078 / 0.542 & 0.077 / 0.504 & 0.151 / 0.685 & 0.196 / 0.712 & 0.177 / 0.563 & 0.068 / 0.585 & 0.070 / 0.682 \\
LLM-SAR & GPT5.5-SAR + knowledge & 0.074 / 0.565 & 0.078 / 0.538 & 0.172 / 0.717 & 0.192 / 0.718 & 0.175 / 0.559 & 0.075 / 0.606 & 0.071 / 0.680 \\
LLM-SAR & Opus4.7-SAR & 0.079 / 0.547 & 0.095 / 0.503 & 0.147 / 0.674 & \underline{0.218} / 0.706 & 0.197 / 0.573 & 0.104 / 0.643 & 0.134 / 0.699 \\
LLM-SAR & Opus4.7-SAR + knowledge & 0.074 / 0.535 & 0.092 / 0.555 & 0.191 / 0.717 & \textbf{\underline{0.226}} / 0.728 & 0.206 / 0.589 & 0.103 / 0.649 & 0.099 / 0.689 \\
\bottomrule
\end{tabular}
}
\end{table*}

%% file: tables/table_structure_toxicity_sr_prauc_rocauc_model_matrix.tex
\begin{table*}[t]
\centering
\caption{Structure separated CV model-by-endpoint matrix for Tox21 stress response assays. Each endpoint entry reports PR-AUC / ROC-AUC. Rankings are computed independently for each metric within each endpoint: the first ranked value is bold and underlined, and the second and third ranked values are underlined.}
\label{tab:table-structure-toxicity-sr-prauc-rocauc-model-matrix}
\tiny
\setlength{\tabcolsep}{2pt}
\resizebox{\textwidth}{!}{%
\begin{tabular}{llccccc}
\toprule
Family & Model & SR-ARE & SR-ATAD5 & SR-HSE & SR-MMP & SR-p53 \\
\midrule
ML & ExtraTrees(ECFP4) & 0.371 / 0.720 & 0.146 / 0.734 & 0.194 / 0.700 & 0.444 / 0.769 & \underline{0.228} / 0.762 \\
ML & ExtraTrees(ECFP6) & 0.362 / 0.722 & \underline{0.155} / 0.741 & 0.179 / 0.691 & 0.446 / 0.780 & 0.209 / 0.745 \\
ML & ExtraTrees(RDKit desc.) & \textbf{\underline{0.443}} / \textbf{\underline{0.778}} & \textbf{\underline{0.178}} / 0.771 & \textbf{\underline{0.301}} / \underline{0.768} & \underline{0.536} / \underline{0.843} & 0.215 / 0.777 \\
ML & ExtraTrees(MACCS) & 0.365 / 0.721 & 0.137 / 0.719 & 0.221 / 0.709 & 0.445 / 0.760 & \underline{0.228} / 0.771 \\
ML & RF(ECFP4) & 0.365 / 0.718 & 0.150 / 0.763 & 0.202 / 0.707 & 0.441 / 0.763 & 0.220 / 0.765 \\
ML & RF(ECFP6) & 0.365 / 0.719 & 0.154 / 0.747 & 0.188 / 0.701 & 0.444 / 0.777 & 0.206 / 0.739 \\
ML & RF(RDKit desc.) & \underline{0.422} / \underline{0.765} & \underline{0.154} / 0.766 & \underline{0.281} / 0.756 & 0.534 / 0.833 & 0.210 / 0.773 \\
ML & RF(MACCS) & 0.377 / 0.729 & 0.132 / 0.715 & 0.225 / 0.713 & 0.457 / 0.767 & 0.222 / 0.776 \\
ML & GBDT(ECFP4) & 0.306 / 0.666 & 0.128 / 0.717 & 0.159 / 0.635 & 0.389 / 0.732 & 0.159 / 0.697 \\
ML & GBDT(ECFP6) & 0.305 / 0.649 & 0.111 / 0.670 & 0.136 / 0.608 & 0.383 / 0.713 & 0.152 / 0.667 \\
ML & GBDT(RDKit desc.) & 0.391 / 0.735 & 0.142 / 0.690 & 0.274 / 0.737 & \underline{0.548} / 0.839 & 0.212 / 0.762 \\
ML & GBDT(MACCS) & 0.337 / 0.698 & 0.127 / 0.633 & 0.232 / 0.700 & 0.390 / 0.719 & 0.191 / 0.740 \\
ML & LR(ECFP4) & 0.263 / 0.615 & 0.098 / 0.635 & 0.117 / 0.548 & 0.291 / 0.654 & 0.146 / 0.641 \\
ML & LR(ECFP6) & 0.227 / 0.581 & 0.102 / 0.675 & 0.088 / 0.564 & 0.304 / 0.677 & 0.134 / 0.633 \\
ML & LR(RDKit desc.) & 0.334 / 0.700 & 0.099 / 0.697 & 0.207 / 0.731 & 0.403 / 0.744 & 0.172 / 0.731 \\
ML & LR(MACCS) & 0.289 / 0.658 & 0.095 / 0.640 & 0.187 / 0.662 & 0.344 / 0.716 & 0.150 / 0.736 \\
\midrule
GNN & Ligandformer & 0.385 / 0.739 & 0.149 / 0.781 & \underline{0.284} / \underline{0.758} & 0.518 / \underline{0.842} & 0.225 / \underline{0.789} \\
GNN & GAT & \underline{0.410} / 0.745 & 0.147 / \underline{0.793} & 0.249 / 0.720 & 0.525 / 0.817 & 0.219 / \underline{0.785} \\
GNN & GCN & 0.371 / 0.729 & 0.152 / \textbf{\underline{0.795}} & 0.229 / 0.738 & 0.492 / 0.809 & 0.191 / 0.775 \\
GNN & GIN & 0.388 / 0.736 & 0.146 / \underline{0.792} & 0.268 / \textbf{\underline{0.770}} & 0.521 / 0.832 & \textbf{\underline{0.245}} / \textbf{\underline{0.806}} \\
\midrule
Sequence & ChemBERTa & 0.335 / 0.718 & 0.112 / 0.714 & 0.185 / 0.722 & 0.470 / 0.815 & 0.169 / 0.730 \\
Sequence & ChemBERTa2 & 0.392 / \underline{0.758} & 0.145 / 0.769 & 0.253 / 0.757 & \textbf{\underline{0.565}} / \textbf{\underline{0.857}} & 0.222 / 0.769 \\
Sequence & MoLFormer & 0.402 / 0.757 & 0.142 / 0.734 & 0.166 / 0.710 & 0.505 / 0.831 & 0.207 / 0.777 \\
\midrule
LLM-SAR & GPT5.5-SAR & 0.225 / 0.586 & 0.053 / 0.589 & 0.082 / 0.577 & 0.245 / 0.609 & 0.082 / 0.573 \\
LLM-SAR & GPT5.5-SAR + knowledge & 0.281 / 0.683 & 0.057 / 0.603 & 0.099 / 0.608 & 0.317 / 0.720 & 0.101 / 0.631 \\
LLM-SAR & Opus4.7-SAR & 0.211 / 0.543 & 0.070 / 0.632 & 0.132 / 0.657 & 0.267 / 0.695 & 0.121 / 0.593 \\
LLM-SAR & Opus4.7-SAR + knowledge & 0.314 / 0.697 & 0.071 / 0.635 & 0.141 / 0.657 & 0.349 / 0.761 & 0.103 / 0.627 \\
\bottomrule
\end{tabular}
}
\end{table*}

%% file: tables/table_structure_bioactivity_prauc_rocauc_model_matrix.tex
\begin{table*}[t]
\centering
\caption{Structure separated CV model-by-endpoint matrix for bioactivity-related classification endpoints. Each endpoint entry reports PR-AUC / ROC-AUC. Rankings are computed independently for each metric within each endpoint: the first ranked value is bold and underlined, and the second and third ranked values are underlined.}
\label{tab:table-structure-bioactivity-prauc-rocauc-model-matrix}
\tiny
\setlength{\tabcolsep}{2pt}
\resizebox{\textwidth}{!}{%
\begin{tabular}{llccc}
\toprule
Family & Model & EGFR & anti-TB H37Rv & antimalaria Pf. 3D7/Dd2 \\
\midrule
ML & ExtraTrees(ECFP4) & \underline{0.953} / 0.754 & \textbf{\underline{0.419}} / 0.722 & 0.827 / 0.827 \\
ML & ExtraTrees(ECFP6) & \underline{0.954} / 0.789 & 0.403 / 0.707 & 0.819 / 0.825 \\
ML & ExtraTrees(RDKit desc.) & 0.929 / 0.691 & 0.384 / 0.721 & 0.836 / 0.849 \\
ML & ExtraTrees(MACCS) & 0.926 / 0.699 & 0.341 / 0.679 & 0.785 / 0.800 \\
ML & RF(ECFP4) & 0.950 / 0.727 & \underline{0.417} / 0.721 & 0.827 / 0.826 \\
ML & RF(ECFP6) & \textbf{\underline{0.955}} / \underline{0.801} & \underline{0.412} / \underline{0.723} & 0.807 / 0.807 \\
ML & RF(RDKit desc.) & 0.926 / 0.688 & 0.378 / 0.715 & 0.822 / 0.831 \\
ML & RF(MACCS) & 0.927 / 0.728 & 0.355 / 0.682 & 0.776 / 0.794 \\
ML & GBDT(ECFP4) & 0.932 / 0.687 & 0.350 / 0.686 & 0.760 / 0.779 \\
ML & GBDT(ECFP6) & 0.933 / 0.674 & 0.364 / 0.694 & 0.743 / 0.778 \\
ML & GBDT(RDKit desc.) & 0.884 / 0.657 & 0.320 / 0.666 & 0.833 / 0.845 \\
ML & GBDT(MACCS) & 0.881 / 0.671 & 0.281 / 0.668 & 0.738 / 0.771 \\
ML & LR(ECFP4) & 0.891 / 0.659 & 0.241 / 0.654 & 0.766 / 0.780 \\
ML & LR(ECFP6) & 0.898 / 0.644 & 0.237 / 0.638 & 0.741 / 0.736 \\
ML & LR(RDKit desc.) & 0.886 / 0.588 & 0.224 / 0.626 & 0.785 / 0.806 \\
ML & LR(MACCS) & 0.881 / 0.697 & 0.237 / 0.631 & 0.679 / 0.718 \\
\midrule
GNN & Ligandformer & 0.941 / \textbf{\underline{0.819}} & 0.380 / 0.716 & \underline{0.845} / \textbf{\underline{0.880}} \\
GNN & GAT & 0.941 / 0.775 & 0.374 / \textbf{\underline{0.728}} & 0.823 / 0.845 \\
GNN & GCN & 0.938 / 0.773 & 0.357 / 0.705 & 0.806 / 0.828 \\
GNN & GIN & 0.942 / \underline{0.814} & 0.366 / \underline{0.726} & \underline{0.845} / \underline{0.868} \\
\midrule
Sequence & ChemBERTa & 0.921 / 0.773 & 0.323 / 0.675 & 0.795 / 0.792 \\
Sequence & ChemBERTa2 & 0.919 / 0.683 & 0.353 / 0.709 & 0.833 / \underline{0.864} \\
Sequence & MoLFormer & 0.936 / 0.760 & 0.386 / 0.708 & \textbf{\underline{0.847}} / 0.863 \\
\midrule
LLM-SAR & GPT5.5-SAR & 0.878 / 0.580 & 0.158 / 0.489 & 0.647 / 0.627 \\
LLM-SAR & GPT5.5-SAR + knowledge & 0.882 / 0.585 & 0.144 / 0.495 & 0.651 / 0.647 \\
LLM-SAR & Opus4.7-SAR & 0.879 / 0.598 & 0.181 / 0.484 & 0.650 / 0.664 \\
LLM-SAR & Opus4.7-SAR + knowledge & 0.903 / 0.580 & 0.192 / 0.589 & 0.658 / 0.703 \\
\bottomrule
\end{tabular}
}
\end{table*}

%% file: tables/table_split_realism_drops.tex
\begin{table}[H]
\centering
\small
\caption{Split realism drop in family best PR-AUC, by endpoint class and model family. Each entry reports the mean family best PR-AUC across endpoint level values in the corresponding group. $\Delta$(random $\rightarrow$ structure) is the random CV mean minus the structure separated CV mean for the same family; larger values indicate larger sensitivity to chemical separation between train and test folds.}
\label{tab:split-realism-drops}
\resizebox{\linewidth}{!}{%
\begin{tabular}{llrrrr}
\toprule
Endpoint class & Family & Random & Murcko & Structure separated & $\Delta$(random $\rightarrow$ structure) \\
\midrule
ADME related & ML & 0.934 & 0.893 & 0.779 & 0.156 \\
 & GNN & 0.933 & 0.904 & 0.816 & 0.117 \\
 & Sequence & 0.943 & 0.899 & 0.804 & 0.139 \\
 & LLM-SAR & 0.822 & 0.809 & 0.710 & 0.111 \\
\midrule
Toxicity related & ML & 0.635 & 0.563 & 0.442 & 0.193 \\
 & GNN & 0.594 & 0.544 & 0.438 & 0.157 \\
 & Sequence & 0.603 & 0.551 & 0.437 & 0.167 \\
 & LLM-SAR & 0.381 & 0.387 & 0.329 & 0.052 \\
\midrule
Bioactivity related & ML & 0.856 & 0.818 & 0.737 & 0.119 \\
 & GNN & 0.839 & 0.801 & 0.722 & 0.117 \\
 & Sequence & 0.834 & 0.801 & 0.723 & 0.112 \\
 & LLM-SAR & 0.647 & 0.713 & 0.584 & 0.063 \\
\bottomrule
\end{tabular}
}
\end{table}

%% file: tables/table_topk_best_ef1.tex
\begin{table}[t]
\centering
\small
\caption{Top K enrichment leaders by endpoint class and model family. EF@1\%, EF@5\% and EF@10\% summarise the family best Top K rows across the split and task entries in each class. ``Best EF@1\%'' and ``Best model'' identify the strongest very early enrichment result in each family and class combination.}
\label{tab:topk-best-ef1}
\resizebox{\linewidth}{!}{%
\begin{tabular}{lllrrrrl}
\toprule
Endpoint class & Family & n & Median EF@1\% & Median EF@5\% & Median EF@10\% & Best EF@1\% & Best model \\
\midrule
ADME-related & ML & 9 & 1.714 & 1.310 & 1.294 & 2.718 & ML:ecfp4:rf \\
 & GNN & 12 & 1.513 & 1.300 & 1.314 & 2.700 & GNN:gat \\
 & Sequence & 12 & 1.311 & 1.305 & 1.283 & 2.553 & Sequence:chemberta \\
\midrule
Toxicity-related & ML & 47 & 9.170 & 5.254 & 3.955 & 27.302 & ML:ecfp6:gbdt \\
 & GNN & 53 & 7.329 & 4.955 & 4.013 & 25.249 & GNN:gcn \\
 & Sequence & 53 & 5.951 & 4.533 & 3.852 & 26.078 & Sequence:chemberta2 \\
\midrule
Bioactivity-related & ML & 7 & 5.474 & 4.163 & 3.410 & 7.110 & ML:ecfp4:gbdt \\
 & GNN & 10 & 2.267 & 2.235 & 2.201 & 6.908 & GNN:gin \\
 & Sequence & 10 & 2.286 & 2.199 & 2.148 & 6.858 & Sequence:chemberta2 \\
\bottomrule
\end{tabular}
}
\end{table}

%% file: tables/table_data_regime_best.tex
\begin{table}[H]
\centering
\small
\caption{Best model at each training data regime for the focused data regime analysis. ``Best'' is the model with the highest fold mean PR-AUC across all three model families at the given training size for the given task; family follows that winning model. Pretrained sequence and classical ML models trade the lead within fold level uncertainty for AMES and Tox21 SR-MMP, while classical ML wins every entry for anti-TB H37Rv and dominates EGFR kinase inhibition once training size reaches at least 1{,}000.}
\label{tab:data-regime-best}
\begin{tabular}{llllll}
\toprule
Task & Training size & Family & Model & PR-AUC & ROC-AUC \\
\midrule
AMES & 500 & Sequence & MoLFormer & 0.759 & 0.742 \\
AMES & 1000 & Sequence & MoLFormer & 0.772 & 0.752 \\
AMES & 3000 & ML & ExtraTrees(RDKit desc.) & 0.797 & 0.767 \\
AMES & full & Sequence & MoLFormer & 0.820 & 0.787 \\
Tox21 SR-MMP & 500 & ML & ExtraTrees(RDKit desc.) & 0.450 & 0.797 \\
Tox21 SR-MMP & 1000 & ML & ExtraTrees(RDKit desc.) & 0.503 & 0.827 \\
Tox21 SR-MMP & 3000 & Sequence & ChemBERTa2 & 0.530 & 0.844 \\
Tox21 SR-MMP & full & Sequence & ChemBERTa2 & 0.541 & 0.854 \\
EGFR kinase inhibition (IC50) & 500 & Sequence & MoLFormer & 0.921 & 0.726 \\
EGFR kinase inhibition (IC50) & 1000 & ML & RF(ECFP4) & 0.937 & 0.706 \\
EGFR kinase inhibition (IC50) & 3000 & ML & RF(ECFP4) & 0.943 & 0.698 \\
EGFR kinase inhibition (IC50) & full & ML & RF(ECFP4) & 0.953 & 0.767 \\
anti-TB H37Rv & 500 & ML & RF(ECFP4) & 0.250 & 0.601 \\
anti-TB H37Rv & 1000 & ML & RF(ECFP4) & 0.261 & 0.616 \\
anti-TB H37Rv & 3000 & ML & RF(ECFP4) & 0.344 & 0.675 \\
anti-TB H37Rv & 10000 & ML & RF(ECFP4) & 0.409 & 0.712 \\
anti-TB H37Rv & full & ML & RF(ECFP4) & 0.416 & 0.719 \\
\bottomrule
\end{tabular}
\end{table}

%% file: tables/table_data_regime_first_crossover.tex
\begin{table}[H]
\centering
\small
\caption{Data regime first crossover summary. ``Crossover size'' is the smallest training size at which a non-ML family achieves the best PR-AUC on the task; ``none up to largest available'' indicates that classical ML remains the best family at every recorded size. The crossover and largest size columns report the winning family, the winning model and its PR-AUC.}
\label{tab:data-regime-first-crossover}
\resizebox{\linewidth}{!}{%
\begin{tabular}{lllllrllr}
\toprule
Task & Crossover size & Crossover family & Crossover model & Crossover PR-AUC & Largest size & Largest family & Largest model & Largest PR-AUC \\
\midrule
AMES & 500 & Sequence & Sequence:molformer & 0.759 & full & Sequence & Sequence:molformer & 0.820 \\
Tox21 SR-MMP & 3000 & Sequence & Sequence:chemberta2 & 0.530 & full & Sequence & Sequence:chemberta2 & 0.541 \\
EGFR kinase IC50 & 500 & Sequence & Sequence:molformer & 0.921 & full & ML & ML:ecfp4:rf & 0.953 \\
anti-TB H37Rv & none up to largest available & - & - & - & full & ML & ML:ecfp4:rf & 0.416 \\
\bottomrule
\end{tabular}
}
\end{table}

%% file: tables/table_prevalence_leaders.tex
\begin{table}[H]
\centering
\small
\caption{External pharmacology and liability primary metric leaders, with prevalence baselines and prevalence adjusted lift. Lift over prevalence is $(\mathrm{PR\text{-}AUC} - p) / (1 - p)$, where $p$ is the held-out positive rate. Lift 1.00 corresponds to a perfect ranker; lift 0.00 corresponds to a model that does no better than the prevalence prior. Endpoint class / subtype labels follow Table~\ref{tab:endpoint-classes}.}
\label{tab:prevalence-leaders}
\resizebox{\linewidth}{!}{%
\begin{tabular}{llllllrr}
\toprule
Endpoint class / subtype & Task & Split & Winner family & Winner model & Primary metric & Prevalence & Lift over prevalence \\
\midrule
Toxicity-related / Off-target CNS safety liability & DRD2 receptor binding (Ki) & Murcko scaffold 5-fold & ML & ExtraTrees(ECFP6) & 0.987 & 89.0\% & 0.883 \\
Toxicity-related / Off-target CNS safety liability & DRD2 receptor binding (Ki) & Random 5-fold & ML & RF(ECFP6) & 0.993 & 89.0\% & 0.939 \\
Toxicity-related / Off-target CNS safety liability & DRD2 receptor binding (Ki) & Structure separated 5-fold & GNN & Ligandformer & 0.973 & 89.0\% & 0.752 \\
Bioactivity-related / Target-based bioactivity & EGFR kinase inhibition (IC50) & Murcko scaffold 5-fold & ML & ExtraTrees(ECFP6) & 0.983 & 79.0\% & 0.917 \\
Bioactivity-related / Target-based bioactivity & EGFR kinase inhibition (IC50) & Random 5-fold & ML & RF(ECFP6) & 0.992 & 79.0\% & 0.960 \\
Bioactivity-related / Target-based bioactivity & EGFR kinase inhibition (IC50) & Structure separated 5-fold & ML & RF(ECFP6) & 0.955 & 79.0\% & 0.788 \\
ADME-related / Metabolism / DDI liability & PXR/NR1I2 activation & Murcko scaffold 5-fold & Sequence & MoLFormer & 0.970 & 77.7\% & 0.865 \\
ADME-related / Metabolism / DDI liability & PXR/NR1I2 activation & Random 5-fold & Sequence & MoLFormer & 0.981 & 77.7\% & 0.915 \\
ADME-related / Metabolism / DDI liability & PXR/NR1I2 activation & Structure separated 5-fold & GNN & Ligandformer & 0.883 & 77.7\% & 0.476 \\
Toxicity-related / Cardiotoxicity / safety pharmacology & hERG/KCNH2 inhibition (IC50) & Murcko scaffold 5-fold & ML & ExtraTrees(RDKit desc.) & 0.812 & 21.8\% & 0.760 \\
Toxicity-related / Cardiotoxicity / safety pharmacology & hERG/KCNH2 inhibition (IC50) & Random 5-fold & ML & ExtraTrees(RDKit desc.) & 0.892 & 21.8\% & 0.861 \\
Toxicity-related / Cardiotoxicity / safety pharmacology & hERG/KCNH2 inhibition (IC50) & Structure separated 5-fold & Sequence & MoLFormer & 0.719 & 21.8\% & 0.641 \\
\bottomrule
\end{tabular}
}
\end{table}

%% file: tables/table_paired_bootstrap_headlines.tex
\begin{table}[H]
\centering
\small
\caption{Paired bootstrap significance summary across the family level and model level. Counts show paired comparisons whose confidence interval excludes zero after Benjamini and Hochberg FDR correction at $q = 0.05$.}
\label{tab:paired-bootstrap-headlines}
\begin{tabular}{llrrrr}
\toprule
Level & Metric & Tests & BH-FDR significant & Fraction & Median paired molecules \\
\midrule
Family level & PR-AUC & 414 & 306 & 0.739 & 6,549 \\
Family level & ROC-AUC & 414 & 307 & 0.742 & 6,549 \\
Family level & Pearson & 54 & 51 & 0.944 & 4,200 \\
Family level & MAE & 54 & 52 & 0.963 & 4,200 \\
Model level & PR-AUC & 207 & 32 & 0.155 & 6,549 \\
Model level & ROC-AUC & 207 & 31 & 0.150 & 6,549 \\
Model level & Pearson & 27 & 13 & 0.481 & 4,200 \\
Model level & MAE & 27 & 17 & 0.630 & 4,200 \\
\bottomrule
\end{tabular}
\end{table}

%% file: tables/table_calibration_best_by_class.tex
\begin{table}[t]
\centering
\small
\caption{Family best calibration by endpoint class. ``Median'' summarises the best PR-AUC model per family per split and task entry across the class. Lower Brier and 10 bin Expected Calibration Error (ECE) are better. LLM-SAR is omitted because per molecule probability scores are not produced by the rule application protocol.}
\label{tab:calibration-best-by-class}
\resizebox{\linewidth}{!}{%
\begin{tabular}{lllrrrr}
\toprule
Endpoint class & Family & n & Median Brier & Min Brier & Median ECE & Max ECE \\
\midrule
ADME-related & ML & 9 & 0.125 & 0.081 & 0.051 & 0.128 \\
 & GNN & 12 & 0.154 & 0.101 & 0.121 & 0.200 \\
 & Sequence & 12 & 0.141 & 0.099 & 0.104 & 0.152 \\
\midrule
Toxicity-related & ML & 47 & 0.070 & 0.018 & 0.022 & 0.168 \\
 & GNN & 53 & 0.099 & 0.036 & 0.094 & 0.269 \\
 & Sequence & 53 & 0.132 & 0.053 & 0.142 & 0.357 \\
\midrule
Bioactivity-related & ML & 7 & 0.092 & 0.073 & 0.044 & 0.080 \\
 & GNN & 10 & 0.127 & 0.069 & 0.113 & 0.167 \\
 & Sequence & 10 & 0.140 & 0.067 & 0.132 & 0.208 \\
\bottomrule
\end{tabular}
}
\end{table}

%% file: tables/table_sar_rule_examples.tex
\begin{table}[H]
\centering
\caption{Examples of SAR rules induced from training folds. Values are fold averaged differences in active rate between compounds with and without the rule feature.}
\label{tab:sar-rules}
\scriptsize
\resizebox{\textwidth}{!}{%
\begin{tabular}{llllr}
\toprule
Task & Representative learned rules & Chemistry readout & Mean $\Delta$ active rate & Mean support \\
\midrule
AMES & nitro; hydrazine; polyaromatic & genotoxic and aromatic alert chemistry & +0.19 to +0.32 & 0.02 to 0.45 \\
Tox21 NR-Aromatase & azole; lipophilic basic amine; high lipophilicity & heme binding and lipophilic endocrine active motifs & +0.10 to +0.12 & 0.09 to 0.18 \\
anti-TB H37Rv & nitroimidazole; perhalo methyl; piperidine like & nitroimidazole and hydrophobic heterocycle pattern signal & +0.11 to +0.28 & 0.03 to 0.13 \\
antimalaria Pf. 3D7/Dd2 & quinoline; secondary aliphatic amine; perhalo methyl & quinoline like and cationic Pf. 3D7/Dd2 activity motifs & +0.25 to +0.37 & 0.21 to 0.30 \\
\bottomrule
\end{tabular}
}
\end{table}